%% file: socioscope.tex
\documentclass[11pt]{article}
\usepackage{fullpage}
\pdfoutput=1
\usepackage{graphicx,amsfonts,amsmath,amsthm}
\usepackage{algorithm2e}
\usepackage{url}

\def\E{\mathbf E}
\def\bff{{\mathbf f}}
\def\bfr{{\mathbf r}}
\def\bfx{\mathbf x}

\def\bfz{\mathbf z}

\def \bP{{\bf P}}

\def \bW{{\bf W}}
\def \bD{{\bf D}}

\def \bL{{\bf L}}
\def \bg{{\bf g}}

\def \bH{{\bf H}}
\def\R{\mathbb R}

\begin{document}

\title{Robust Spatio-Temporal Signal Recovery\\from Noisy Counts in Social Media}
\author{Jun-Ming Xu$^\dagger$ ~~ Aniruddha Bhargava$^*$ ~~ Robert Nowak$^*$ ~~ Xiaojin Zhu$^{\dagger*}$\\
$^\dagger$ Department of Computer Sciences\\
$^*$ Department of Electrical and Computer Engineering\\
University of Wisconsin-Madison\\
\textit{xujm@cs.wisc.edu, aniruddha@wisc.edu, nowak@ece.wisc.edu, jerryzhu@cs.wisc.edu}
}
\date{}
\maketitle
\begin{abstract}
\begin{quote}
Many real-world phenomena can be represented by a spatio-temporal signal: where, when, and how much.
Social media is a tantalizing data source for those who wish to monitor such signals.
Unlike most prior work, we assume that the target phenomenon is known and we are given a method to count its occurrences in social media.
However, counting is plagued by sample bias, incomplete data, and, paradoxically, data scarcity -- issues inadequately addressed by prior work.
We formulate signal recovery as a Poisson point process estimation problem.
We explicitly incorporate human population bias, time delays and spatial distortions, and spatio-temporal regularization into the model to address the noisy count issues. 
We present an efficient optimization algorithm and discuss its theoretical properties.
We show that our model is more accurate than commonly-used baselines.
Finally, we present a case study on wildlife roadkill monitoring, where our model produces qualitatively convincing results.
\end{quote}
\end{abstract}

\renewcommand{\baselinestretch}{0.95}

\input{intro}

\input{sscope}
\input{relatedWork}
\input{toy}
\input{roadkill}
\input{conclusion}

\section{Acknowledgments}
We thank Megan K. Hines from Wildlife Data Integration Network for providing range maps and guidance on wildlife.

\bibliography{socioscope}
\bibliographystyle{plain}
\end{document}

%% file: intro.tex
\section{Introduction}

Many real-world phenomena of interest to science are spatio-temporal in nature.
They can be characterized by a real-valued intensity function $\mathbf{f} \in \R_{\ge 0}$, where the value $f_{s,t}$ quantifies the prevalence of the phenomenon at location $s$ and time $t$.
Examples include wildlife mortality, algal blooms, hail damage, and seismic intensity. 
Direct instrumental sensing of $\mathbf{f}$ is often difficult and expensive.
Social media offers a unique sensing opportunity for such spatio-temporal signals, where users serve the role of ``sensors'' by posting their experiences of a target phenomenon.
For instance, social media users readily post their encounters with dead animals:
\textit{``I saw a dead crow on its back in the middle of the road.''}

There are at least three challenges faced when using human social media users as sensors:
\begin{enumerate}
\item Social media sources are not always reliable and consistent, due to factors including the vagaries of language and the psychology of users.  This makes identifying topics of interest and labeling social media posts extremely challenging.
\item Social media users are not under our control.  In most cases, users cannot be directed or focused or maneuvered as we wish.  The distribution of human users (our sensors) depends on many factors unrelated to the sensing task at hand.
\item Location and time stamps associated with social media posts may be erroneous or missing.  Most posts do not include GPS coordinates, and self-reported locations can be inaccurate or false. Furthermore, there can be random delays between an event of interest and the time of the social media post related to the event.
\end{enumerate}
Most prior work in social media event analysis has focused on the first challenge. Sophisticated natural language processing techniques have been used to identify social media posts relevant to a topic of interest~\cite{Yang1998,Becker2011,Sakaki2010} and advanced machine learning tools have been proposed to discover popular or emerging topics in social media~\cite{Allan2002,Mei2006,Yin2011}.
We discuss the related work in detail in Section~\ref{sec:related}.  

Our work in this paper focuses on the latter two challenges.  We are interested in a specific topic or target phenomenon of interest that is given and fixed beforehand, and we assume that we are also given a (perhaps imperfect) method, such as a trained text classifier, to identify target posts.  The first challenge is relevant here, but is not the focus of our work.  The main concerns of this paper are to deal with the highly non-uniform distribution of human users (sensors), which profoundly affects our capabilities for sensing natural phenomena such as  wildlife mortality, and to cope with the uncertainties in the location and time stamps associated with related social media posts.  The main contribution of the paper is robust methodology for deriving accurate spatiotemporal maps of the target phenomenon in light of these two challenges.

%% file: sscope.tex
\newtheorem{theorem}{Theorem}
\section{The Socioscope}

We propose Socioscope, a probabilistic model that robustly recovers spatiotemporal signals from social media data.
Formally, consider $\mathbf{f}$ defined on discrete spatiotemporal bins.
For example, a bin $(s,t)$ could be a U.S. state $s$ on day $t$, or a county $s$ in hour $t$.
From the first stage we obtain $x_{s,t}$, the count of target social media posts within that bin.
The task is to estimate $f_{s,t}$ from $x_{s,t}$.
A commonly-used estimate is $\widehat{f_{s,t}} = x_{s,t}$ itself.
This estimate can be justified as the maximum likelihood estimate of a Poisson model 
$\mathbf{x} \sim \mbox{Poisson}(\mathbf{f})$.
This idea underlines several emerging systems such as earthquake damage monitoring from Twitter~\cite{Earle2009OMG}.
However, this estimate is unsatisfactory since the counts $x_{s,t}$ can be \emph{noisy}:
as mentioned before, the estimate ignores population bias -- more target posts are generated when and where there are more social media users; 
the location of a target post is frequently inaccurate or missing, making it difficult to assign to the correct bin;
and target posts can be quite sparse even though the total volume of social media is huge.
Socioscope addresses these issues.

For notational simplicity, we often denote our signal of interest by a vector $\bff = (f_1, \ldots, f_n)^\top \in \R_{\ge 0}^n$, where $f_j$ is a non-negative target phenomenon intensity in \emph{source bin} $j=1 \ldots n$.
We will use a wildlife example throughout the section. 
In this example, 
a source bin is a spatiotemporal unit such as ``California, day 1,''
and $f_j$ is the squirrel activity level in that unit. 
The mapping between index $j$ and the aforementioned $(s,t)$ is one-one and will be clear from context.

\subsection{Correcting Human Population Bias}
For now, assume each target post comes with precise location and time meta data.
This allows us to count $x_j$, the number of target posts in bin $j$.
Given $x_j$, it is tempting to use the maximum likelihood estimate $\widehat{f_j}=x_j$ which assumes a simple Poisson model $x_j \sim \mathrm{Poisson}(f_j)$.
However, this model is too naive:
Even if $f_j=f_k$, e.g., the level of squirrel activities is the same in two bins, we would expect $x_j > x_k$ if there are more people in bin $j$ than in bin $k$, simply because more people see the same group of squirrels.

To account for this population bias, 
we define an ``active social media user population intensity'' (loosely called ``human population'' below) $\bg = (g_1, \ldots, g_n)^\top \in \R_{\ge 0}^n$. 
Let $z_j$ be the count of \emph{all} social media posts in bin $j$, the vast majority of which are not about the target phenomenon.
We assume $z_j \sim \mathrm{Poisson}(g_j)$.
Since typically $z_j \gg 0$, the maximum likelihood estimate $\widehat{g_j}=z_j$ is reasonable.

Importantly, we then posit the Poisson model
\begin{equation}
x_j \sim \mathrm{Poisson}(\eta(f_j,g_j)).
\end{equation}
The intensity is defined by a 
\emph{link function} $\eta(f_j,g_j)$.
In this paper, we simply define $\eta(f_j,g_j)=f_j \cdot g_j$ but note that other more sophisticated link functions can be learned from data. 
Given $x_j$ and $z_j$, one can then easily estimate $f_j$ with the plug-in estimator $\widehat{f_j}=x_j / z_j$.

\subsection{Handling Noisy and Incomplete Data}
This would have been the end of the story if we could reliably assign each post to a source bin.
Unfortunately, this is often not the case for social media.
In this paper, we focus on the problem of spatial uncertainty due to noisy or incomplete social media data. 
A prime example of spatial uncertainty is the lack of location meta data in posts from Twitter (called tweets).\footnote{It may be possible to recover occasional location information from the tweet text itself instead of the meta data, but the problem still exists.}
In recent data we collected,
only 3\% of tweets contain the latitude and longitude at which they were created.
Another 47\% contain a valid user self-declared location in his or her profile (e.g., ``New York, NY'').
However, such location does not automatically change while the user travels and thus may not be the true location at which a tweet is posted.
The remaining 50\% do not contain location at all.
Clearly, we cannot reliably assign the latter two kinds of tweets to a spatiotemporal source bin.
\footnote{\label{fn:psf}Another kind of spatiotemporal uncertainty exists in social media even when the local and time meta data of every post is known: social media users may not immediately post right at the spot where a target phenomenon happens.
Instead, there usually is an unknown time delay and spatial shift between the phenomenon and the post generation.
For example, one may not post a squirrel encounter on the road until she arrives at home later; the local and time meta data only reflects tweet-generation at home.
This type of spatiotemporal uncertainty can be addressed by the same source-detector transition model.
}

To address this issue, we borrow an idea from Positron Emission Tomography~\cite{vardi85}.
In particular, we define $m$ \emph{detector bins} which are conceptually distinct from the $n$ source bins.
The idea is that an event originating in some source bin goes through a transition process and ends up in one of the detector bins, where it is detected.
This transition is modeled by 
an $m \times n$ matrix $P$ where 
\begin{equation}
P_{ij}= \mathrm{Pr}(\mbox{detector} \; i \mid \mbox{source} \; j).
\end{equation}
$P$ is column stochastic: $\sum_{i=1}^m P_{ij}=1,\forall j$.
We defer the discussion of our specific $P$ to a case study, but we mention that it is possible to reliably estimate $P$ directly from social media data (more on this later).
Recall the target post intensity at source bin $j$ is $\eta(f_j,g_j)$.
We use the transition matrix to define the target post intensity $h_i$ (note that $h_i$ can itself be viewed as a link function $\tilde{\eta}(\mathbf{f},\mathbf{g})$) at detector bin $i$:
\begin{equation}
h_i = \sum_{j=1}^n P_{ij} \eta(f_j,g_j).
\end{equation}

For the spatial uncertainty that we consider, we create three kinds of detector bins.
For a source bin $j$ such as ``California, day 1,''
the first kind collects target posts on day 1 whose latitude and longitude meta data is in California. 
The second kind collects target posts on day 1 without latitude and longitude meta data, but whose user self-declared profile location is in California. 
The third kind collects target posts on day 1 without any location information.
Note the third kind of detector bin is shared by all other source bins for day 1, such as ``Nevada, day 1,'' too.
Consequently, if we had $n=50T$ source bins corresponding to the 50 US states over $T$ days, there would be $m=(2\times 50+1)T$ detector bins. 

Critically, our observed target counts $\bfx$ are now with respect to the $m$ detector bins instead of the $n$ source bins: $\bfx = (x_1, \ldots, x_m)^\top$.
We will also denote the count sub-vector for the first kind of detector bins by $\bfx^{(1)}$, the second kind $\bfx^{(2)}$, and the third kind $\bfx^{(3)}$.
The same is true for the overall counts $\bfz$.
A trivial approach is to only utilize $\bfx^{(1)}$ and $\bfz^{(1)}$ to arrive at the plug-in estimator 
\begin{equation}
\widehat{f_j}=x^{(1)}_j / z^{(1)}_j.
\label{eq:xoverz}
\end{equation}
As we will show, we can obtain a better estimator by incorporating noisy data $\bfx^{(2)}$ and incomplete data $\bfx^{(3)}$.
$\bfz^{(1)}$ is sufficiently large and we will simply ignore $\bfz^{(2)}$ and $\bfz^{(3)}$.

\subsection{Socioscope: Penalized Poisson Likelihood Model}
We observe target post counts $\bfx=(x_1, \ldots, x_m)$ in the detector bins.
These are modeled as independently Poisson distributed random variables: 
\begin{equation}
x_i \sim \mathrm{Poisson}(h_i), \; \mbox{for } i=1 \ldots m.
\end{equation}
The log likelihood factors as 
\begin{equation}
\ell(\bff) = \log \prod_{i=1}^m \frac{h_i^{x_i} e^{-h_i}}{x_i!} = \sum_{i=1}^m \left(x_i \log h_i - h_i \right) + c,
\label{eq:loglik}
\end{equation}
where $c$ is a constant.
In~\eqref{eq:loglik} we treat $g$ as given.

Target posts may be scarce in some detector bins.
Indeed, we often have zero target posts for the wildlife case study to be discussed later.
This problem can be mitigated by the fact that
many real-world phenomena are spatiotemporally smooth, i.e., ``neighboring'' source bins in space or time tend to have similar intensity.
We therefore adopt a penalized likelihood approach by constructing a graph-based regularizer.
The undirected graph is constructed so that the nodes are the source bins.
Let $\bW$ be the $n \times n$ symmetric non-negative weight matrix.
The edge weights are such that $w_{jk}$ is large if $j$ and $k$ correspond to neighboring bins in space and time.
Since $\bW$ is domain specific, we defer its construction to the case study.

Before discussing the regularizer, we need to perform a change of variables.
Poisson intensity $\bff$ is non-negative, necessitating a constrained optimization problem.
It is more convenient to work with an unconstrained problem.
To this end, we work with the exponential family natural parameters of Poisson.
Specifically, let
\begin{equation}
\theta_j = \log f_j, \; \; \psi_j = \log g_j.
\end{equation}
Our specific link function becomes 
$\eta(\theta_j, \psi_j) = e^{\theta_j + \psi_j}$. 
The detector bin intensities become
$h_i = \sum_{j=1}^n P_{ij} \eta(\theta_j, \psi_j)$.

Our graph-based regularizer applies to $\theta$ directly:
\begin{equation}
\Omega(\theta) = \frac{1}{2} \theta^\top \bL \theta,
\end{equation}
where $\bL$ is the combinatorial graph Laplacian~\cite{chung97spectral}:
$\bL = \bD-\bW$,
and $\bD$ is the diagonal degree matrix with $D_{jj}=\sum_{k=1}^n w_{jk}$.

Finally, Socioscope is the following penalized likelihood optimization problem: 
\begin{equation}
\min_{\theta \in \R^n} - \sum_{i=1}^m \left(x_i \log h_i - h_i \right) + \lambda \Omega(\theta),
\label{eq:opt}
\end{equation}
where $\lambda$ is a positive regularization weight.

\subsection{Optimization}
We solve the Socioscope optimization problem~\eqref{eq:opt} with BFGS, a quasi-Newton method~\cite{nocedal1999numerical}. 
The gradient can be easily computed as
\begin{equation}
\nabla = \lambda \bL \theta - \bH P^\top (\bfr - 1),
\end{equation}
where $\bfr=(r_1 \ldots r_m)$ is a ratio vector with $r_i = x_i / h_i$, and $\bH$ is a diagonal matrix with $\bH_{jj} = \eta(\theta_j, \psi_j)$.

We initialize $\theta$ with the following heuristic.
Given counts $\bfx$ and the transition matrix $P$, we compute the least-squared projection $\eta_0$ to $\| \bfx - P \eta_0 \|_2$.
This projection is easy to compute.
However, $\eta_0$ may contain negative components not suitable for Poisson intensity.
We force positivity by setting $\eta_0 \leftarrow \max(10^{-4}, \eta_0)$ element-wise, where the floor $10^{-4}$ ensures that $\log \eta_0 > -\infty$.
From the definition $\eta(\theta,\psi)=\exp(\theta+\psi)$, we then obtain the initial parameter
\begin{equation}
\theta_0 = \log \eta_0 - \psi. 
\end{equation}

Our optimization is efficient: problems with more than one thousand variables ($n$) are solved in about 15 seconds with fminunc() in Matlab.

\subsection{Parameter Tuning}
The choice of the regularization parameter $\lambda$ has a profound effect on the smoothness of the estimates.  
It may be possible to select these parameters based on prior knowledge in certain problems, but for our experiments we select these parameters using a cross-validation (CV) procedure, which gives us a fully data-based and objective approach to regularization.  

CV is quite simple to implement in the Poisson setting.  
A hold-out set of data can be constructed by simply sub-sampling events from the total observation uniformly at random.  
This produces a partial data set of a subset of the counts that follows precisely the same distribution as the whole set, modulo a decrease in the total intensity per the level
of subsampling.  
The complement of the hold-out set is what remains of the full dataset, and we will call this the training set. 
The hold-out set is taken to be a specific fraction of the total.  
For theoretical reasons beyond the scope of this paper, we do not recommend leave-one-out CV~\cite{van2003unified,cornec2010concentration}.

CV is implemented by generating a number of random splits of this type (we can generate as many as we wish), and for each split we run the optimization algorithm above on the training set for a range of values of $\lambda$. 
Then compute the (unregularized) value of the log-likelihood on the hold-out set.  
This provides us with an estimate of the log-likelihood for each setting of $\lambda$.  
We simply select the setting that maximizes the estimated log-likelihood.

\subsection{Theoretical Considerations}

The natural measure of signal-to-noise in this problem is the number of counts in each bin.  The higher the counts, the more stable and ``less noisy'' our estimators will be.  Indeed, if we directly observe $x_i \sim \mbox{Poisson}(h_i)$, then the normalized error $\E[(\frac{x_i-h_i}{h_i})^2] = h_i^{-1} \approx x_i^{-1}$.
So larger counts, due to larger underlying intensities, lead to small errors on a relative scale.  However, the accuracy of our recovery also depends on the regularity of the underlying function $f$.  If it is very smooth, for example a constant function, then the error would be inversely proportional to the total number of counts, not the number in each individual bin.  This is because in the extreme smooth case,  $f$ is determined by a single constant.

To give some insight into dependence of the estimate on the total number of counts, suppose that $f$ is the underlying continuous intensity function of interest.  Furthermore, let $f$ be a H{\"o}lder $\alpha$-smooth function.  The parameter $\alpha$ is related to the number of continuous derivatives $f$ has.
Larger values of $\alpha$ correspond to smoother functions.  Such a model is reasonable for the application at hand, as discussed in our motivation for regularization above. We recall the following minimax lower bound, which follows from the results in \cite{donoho,willett}.
\begin{theorem}  Let $f$ be a H{\"o}lder $\alpha$-smooth $d$-dimensional intensity function and suppose we observe $N$ events from the distribution $\mbox{Poisson}(f)$.  Then there exists a constant $C_\alpha>0$ such that
$$\inf_{\widehat{f}} \sup_f  \frac{\E[\|\widehat{f}-f\|_1^2]}{\|f\|_1^2} \ \geq \ C_\alpha N^{\frac{-2\alpha}{2\alpha+d}} \ ,$$
\end{theorem}
\noindent where the infimum is over all possible estimators.  The error is measured with the $1$-norm, rather than two norm, which is a more appropriate and natural norm in density and intensity estimation.  The theorem tells us that no estimator can achieve a faster rate of error decay than the bound above.  There exist many types of estimators that nearly achieve this bound (e.g., to within a log factor), and with more work it is possible to show that our regularized estimators, with adaptively chosen bin sizes and appropriate regularization parameter settings, could also nearly achieve this rate.  For the purposes of this discussion, the lower bound, which certainly applies to our situation, will suffice. 

For example, consider just two spatial dimensions ($d=2$) and $\alpha=1$ which corresponds to Lipschitz smooth functions, a very mild regularity assumption.    Then the bound says that the error is proportional to $N^{-1/2}$.   This gives useful insight into the minimal data requirements of our methods.   It tells us, for example, that if we want to reduce the error of the estimator by a factor of say $2$, then the total number of counts must be increased by a factor of $4$. If the smoothness $\alpha$ is very large, then doubling the counts can halve the error.  The message is simple.  More events and higher counts will provide more accurate estimates.

%% file: relatedWork.tex
\section{Related Work}
\label{sec:related}

To our knowledge, there is no comparable prior work that focuses on robust single recovery from social media (i.e., the ``second stage'' as we mentioned in the introduction).
However, there has been considerable related work on the first stage, which we summarize below.

Topic detection and tracking (TDT) aims at identifying emerging topics from text stream and grouping documents based on their topics.
The early work in this direction began with news text streamed from newswire and transcribed from other media~\cite{Allan2002}.
Recent research focused on user-generated content on the web and on the spatio-temporal variation of topics.
Latent Dirichlet Allocation (LDA)~\cite{Blei2003, griffiths2004} is a popular unsupervised method to detect topics.
Mei \emph{et~al.}~\cite{Mei2006} extended LDA by taking spatio-temporal context into account to identify subtopics from weblogs.
They analyzed the spatio-temporal pattern of topic $\theta$ by $p(time|\theta, location)$ and $p(location|\theta, time)$, 
and showed that documents created from the same spatio-temporal context tend to share topics. 
In the same spirit, Yin \emph{et~al.}~\cite{Yin2011} studied GPS-associated documents, whose coordinates are generated by Gaussian Mixture Model in their generative framework.
Cataldi \emph{et~al.}~\cite{Cataldi2010} proposed a \textit{feature-pivot} method.
They first identified keywords whose occurrences dramatically increase in a specified time interval and then connected the keywords to detect emerging topics. 
Besides text, social network structure also provides important information for detecting community-based topics~\cite{Qamra2006} and user interests~\cite{Lin2010}.

Event detection is highly related to TDT.
Yang \emph{et~al.}~\cite{Yang1998} uses clustering algorithm to identify events from news stream.
Others tried to distinguish posts related to real world event from non-events ones, such as describing daily life or emotions~\cite{Becker2011}.  
Such kind of events were also detected in Flickr photos with meta information~\cite{Chen2009} and Twitter~\cite{weng2011}.
Yet others were interested in events with special characteristics.
Popescu \emph{et~al.}~\cite{Popescu2010,Popescu2011} focused on the detection of controversial events which provoke a public debate in which audience members express opposing opinions. 
Watanabe \emph{et~al.}~\cite{Watanabe2011} studied smaller-scale local-events, such as sales at a supermarket.
Sakaki \emph{et~al.}~\cite{Sakaki2010} monitored Twitter to detect real-time events such as earthquakes and hurricanes.

Another line of related work uses social media as a data source to answer scientific questions~\cite{lazer2009}.
Most previous work studied questions in linguistic, sociology and human interactions. 
For example, Eisenstein \emph{et~al.}~\cite{Eisenstein2010} studied the geographic linguistic variation with geotagged social media. 
Danescu-Niculescu-Mizil \emph{et~al.}~\cite{Mizil2011} studied the psycholinguistic theory of communication accommodation with twitter conversations.
Gupte \emph{et~al.}~\cite{Gupte2011} studied social hierarchy and stratification in online social network.
Crandall \emph{et~al.}~\cite{Crandall2008} and Anagnostopoulos \emph{et~al.}~\cite{Anagnostopoulos2008} tried to understand the social influence through the interaction on social network.

As stated earlier, Socioscope differs from these related work in its focus on robust signal recovery on predefined target phenomena.
The target posts may be generated at a very low, though sustained, rate, and are subject to noise corruption.
The above approaches are unlikely to estimate the underlying intensity accurately.

%% file: toy.tex
\section{A Synthetic Experiment}

We start with a synthetic experiment whose known ground-truth intensity $\bff$ allows us to quantitatively evaluate the effectiveness of Socioscope. 
The synthetic experiment matches the case study in the next section.
There are 48 US continental states plus Washington DC, and $T=24$ hours.
This leads to a total of $n=1176$ source bins, and $m = (2\times 49+1) T = 2376$ detector bins.
The transition matrix $\bP$ is the same as in the case study, to be discussed later.
The overall counts $\bfz$ are obtained from actual Twitter data and $\widehat\bg = \bfz^{(1)}$.

\begin{figure}
\centering
\begin{tabular}{ccc}
\includegraphics[width=.3\columnwidth]{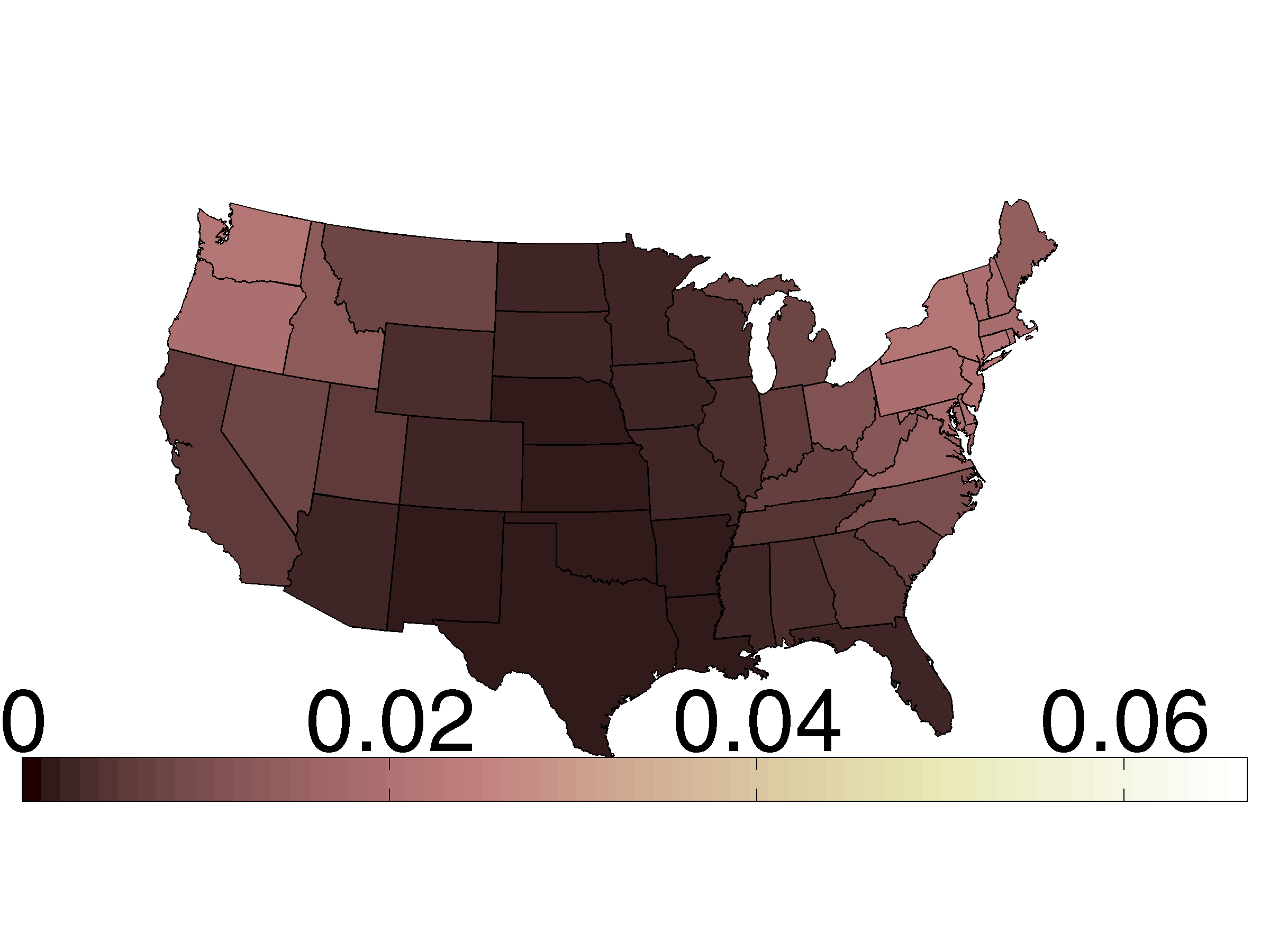} &
\includegraphics[width=.3\columnwidth]{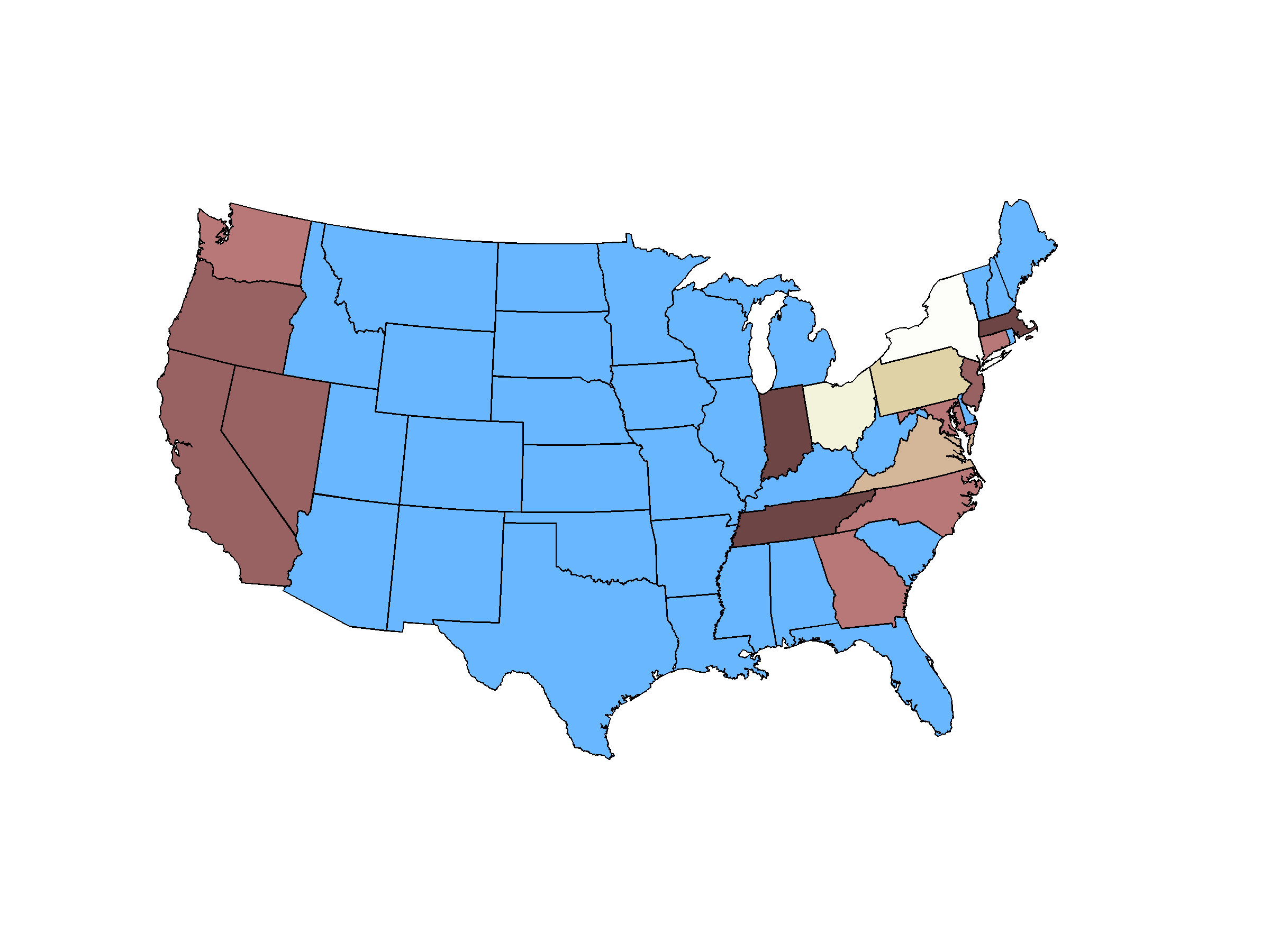} &
\includegraphics[width=.3\columnwidth]{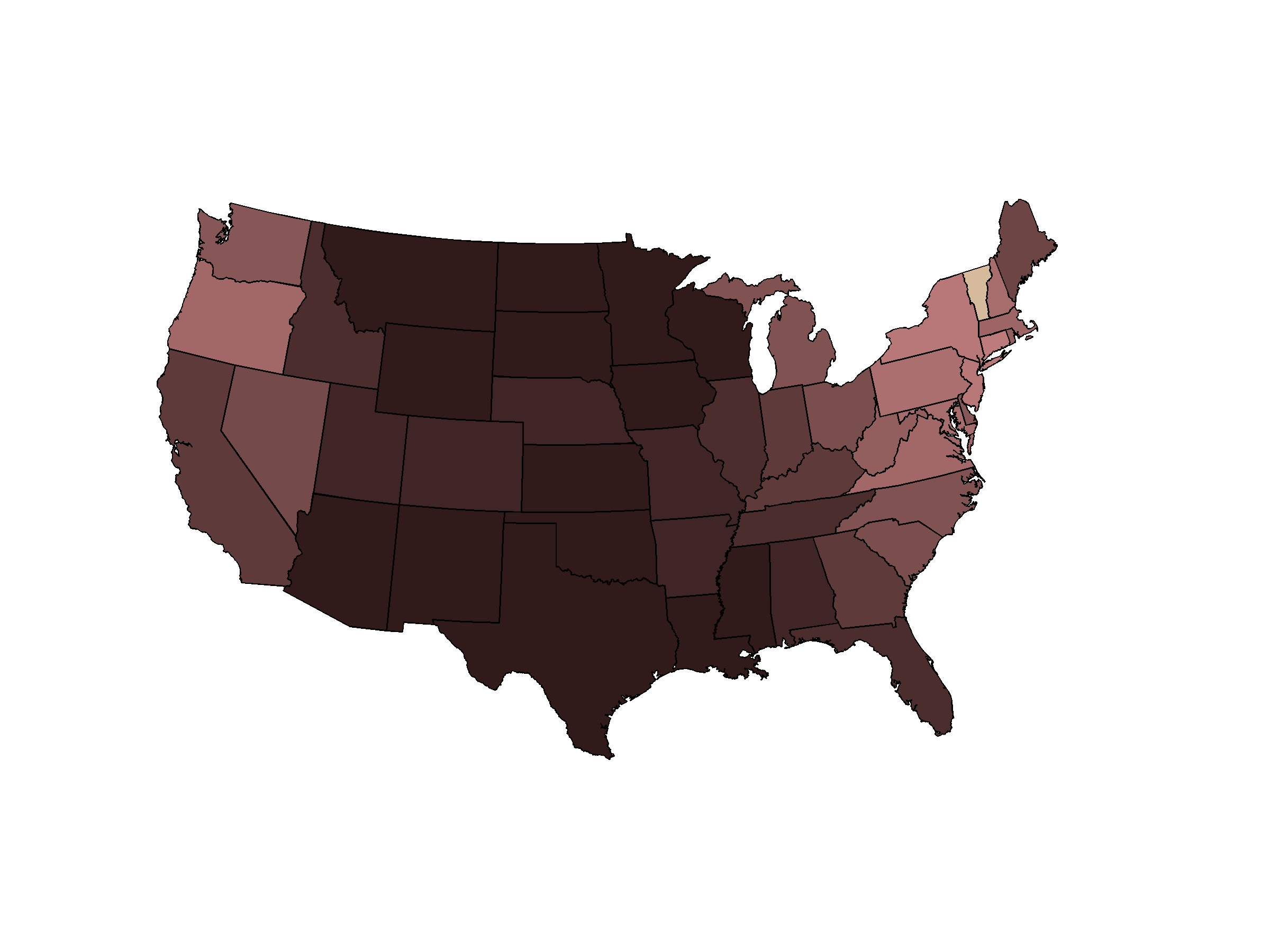} \\
(a) ground-truth $\bff$  & (b) scaled $\bfx^{(1)}$  &  (c) Socioscope\\
\end{tabular}
\caption{The synthetic experiment}
\label{fig:toy}
\vskip -2ex
\end{figure}

We design the ground-truth target signal $\bff$ to be temporally constant but spatially varying.
Figure~\ref{fig:toy}(a) shows the ground-truth $\bff$ spatially.
It is a mixture of two Gaussian distributions discretized at the state level.
The modes are in Washington and New York, respectively.
From $\bP$, $\bff$ and $\bg$, we generate the observed target post counts for each detector bin by a Poisson random number generator: $x_i \sim \mbox{Poisson}(\sum_{j=1}^n P_{i,j} f_j g_j)$, $i=1\ldots m$.
The sum of counts in $\bfx^{(1)}$ is 56, in $\bfx^{(2)}$ 1106, and in $\bfx^{(3)}$ 1030.
Considering the number of bins we have, the data is very sparse.

\begin{table}
\centering
\begin{tabular}{|l|r|}
\hline
(i) scaled $\bfx^{(1)}$ & 14.11\\
(ii) scaled $\bfx^{(1)} / \bfz^{(1)}$ & 46.73 \\
(iii) Socioscope with $\bfx^{(1)}$ & 0.17 \\
(iv) Socioscope with $\bfx^{(1)} + \bfx^{(2)} $ & 1.83 \\
(v) Socioscope with $\bfx^{(1)}$, $\bfx^{(2)}$& 0.16 \\
(vi) \textbf{Socioscope with} $\bfx^{(1)}$, $\bfx^{(2)}$, $\bfx^{(3)}$ & \textbf{0.12} \\
\hline
\end{tabular}
\caption{Relative error of different estimators}
\label{tab:mse}
\vskip -2ex
\end{table}

Given $\bfx, \bP, \bg$, 
We compare the relative error $\|\bff - \hat{\bff}\|^2/ \|\bff\|^2$ of several estimators in Table~\ref{tab:mse}:

(i)
$\hat{\bff} = \bfx^{(1)} / (\epsilon_1 \sum \bfz^{(1)})$,
where $\epsilon_1$ is the fraction of tweets with precise location stamp (discussed later in case study).
Scaling matches it to the other estimators. 
Figure~\ref{fig:toy}(b) shows this simple estimator, aggregated spatially.
It is a poor estimator: besides being non-smooth, it contains 32 ``holes'' (states with zero intensity, colored in blue)  due to data scarcity.
(ii)
$\hat{\bff} = \bfx_j^{(1)}/ (\epsilon_1 \bfz_j^{(1)})$ which naively corrects the population bias as discussed in~\eqref{eq:xoverz}.
It is even worse than the simple estimator, because naive bin-wise correction magnifies the variance in sparse $\bfx^{(1)}$.

(iii) Socioscope with $\bfx^{(1)}$ only.
This simulates the practice of discarding noisy or incomplete data, but regularizing for smoothness.
The relative error was reduced dramatically.

(iv) Same as (iii) but replace the values of $\bfx^{(1)}$ with $\bfx^{(1)} + \bfx^{(2)}$.
This simulates the practice of ignoring the noise in $\bfx^{(2)}$ and pretending it is precise.
The result is worse than (iii), indicating that simply including noisy data may hurt the estimation.

(v) Socioscope with $\bfx^{(1)}$ and $\bfx^{(2)}$ separately, where $\bfx^{(2)}$ is treated as noisy by $P$.
It reduces the relative error further, and demonstrates the benefits of treating noisy data specially.

(vi) Socioscope with the full $\bfx$.
It achieves the lowest relative error among all methods, and is the closest to the ground truth (Figure~\ref{fig:toy}(c)).
Compared to (v), this demonstrates that even counts $\bfx^{(3)}$ without location can also help us to recover $\bff$ better.

%% file: roadkill.tex
\section{Case Study: Roadkill}

We now turn to a real-world task of estimating the spatio-temporal intensity of roadkill for several common wildlife species from Twitter posts.
The study of roadkill has values in ecology, conservation, and transportation safety.

The target phenomenon consists of roadkill events for a specific species within the continental United States during September 22--November 30, 2011.
Our spatio-temporal source bins are state$\times$hour-of-day.
Let $s$ index the 48 continental US states plus District of Columbia.
We aggregate the 10-week study period into 24 hours of a day.
The target counts $\bfx$ are still sparse even with aggregation:
for example, most state-hour combination have zero counts for armadillo and the largest count in $\bfx^{(1)}$ and $\bfx^{(2)}$ is 3.
Therefore, recovering the underlying signal $\bff$ remains a challenge.
Let $t$ index the hours from 1 to 24.
This results in $|s|=49, |t|=24, n=|s||t| = 1176, m=(2|s|+1)|t|=2376$.
We will often index source or detector bins by the subscript $(s,t)$, in addition to $i$ or $j$, below.
The translation should be obvious.

\subsection{Data Preparation}

We chose Twitter as our data source because public tweets can be easily collected through its APIs. 
All tweets include time meta data. However, most tweets do not contain location meta data, as discussed earlier.

\begin{figure}[t!]
\centering
\begin{tabular}{ccc}
\includegraphics[width=.3\columnwidth]{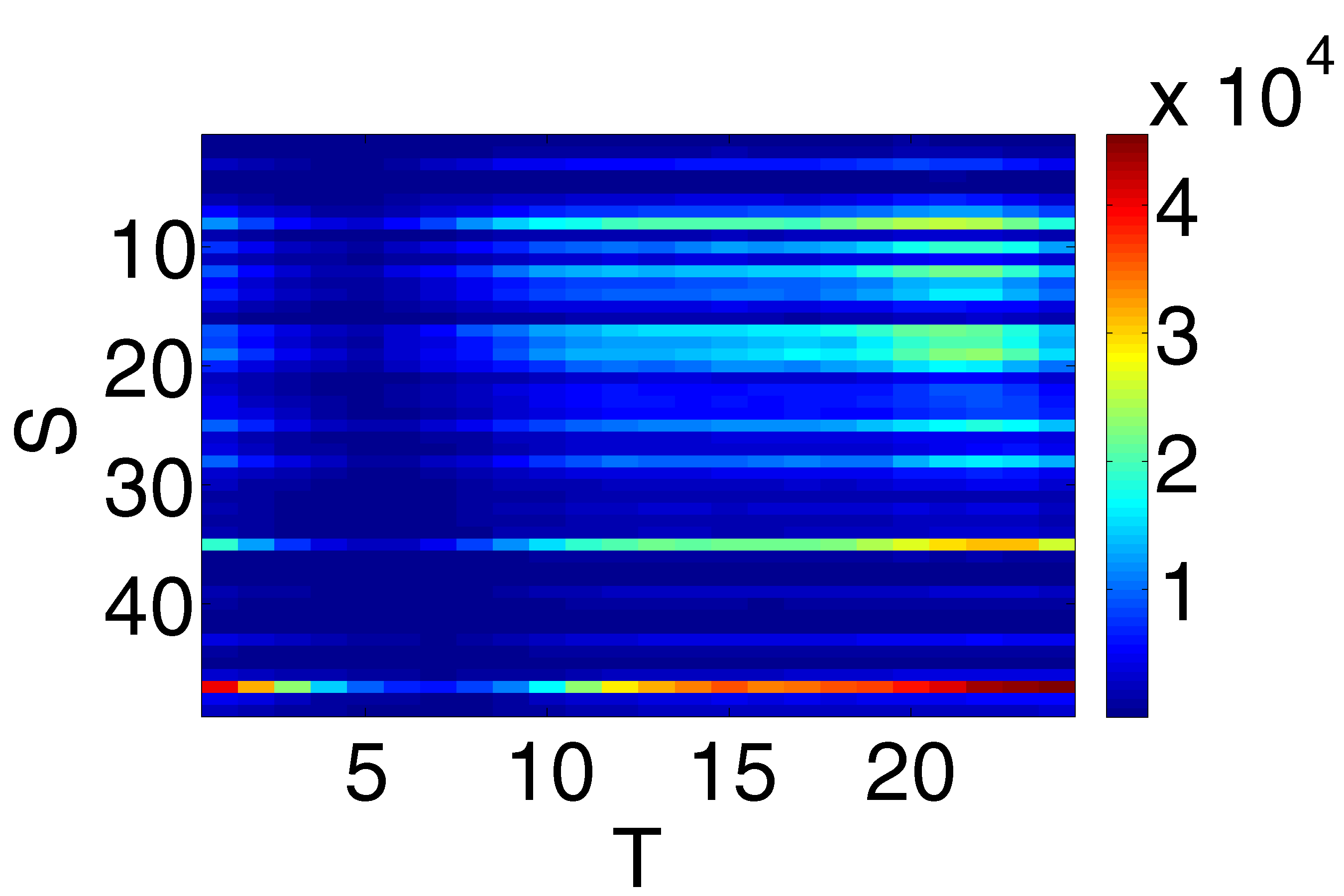} &
\includegraphics[width=.3\columnwidth]{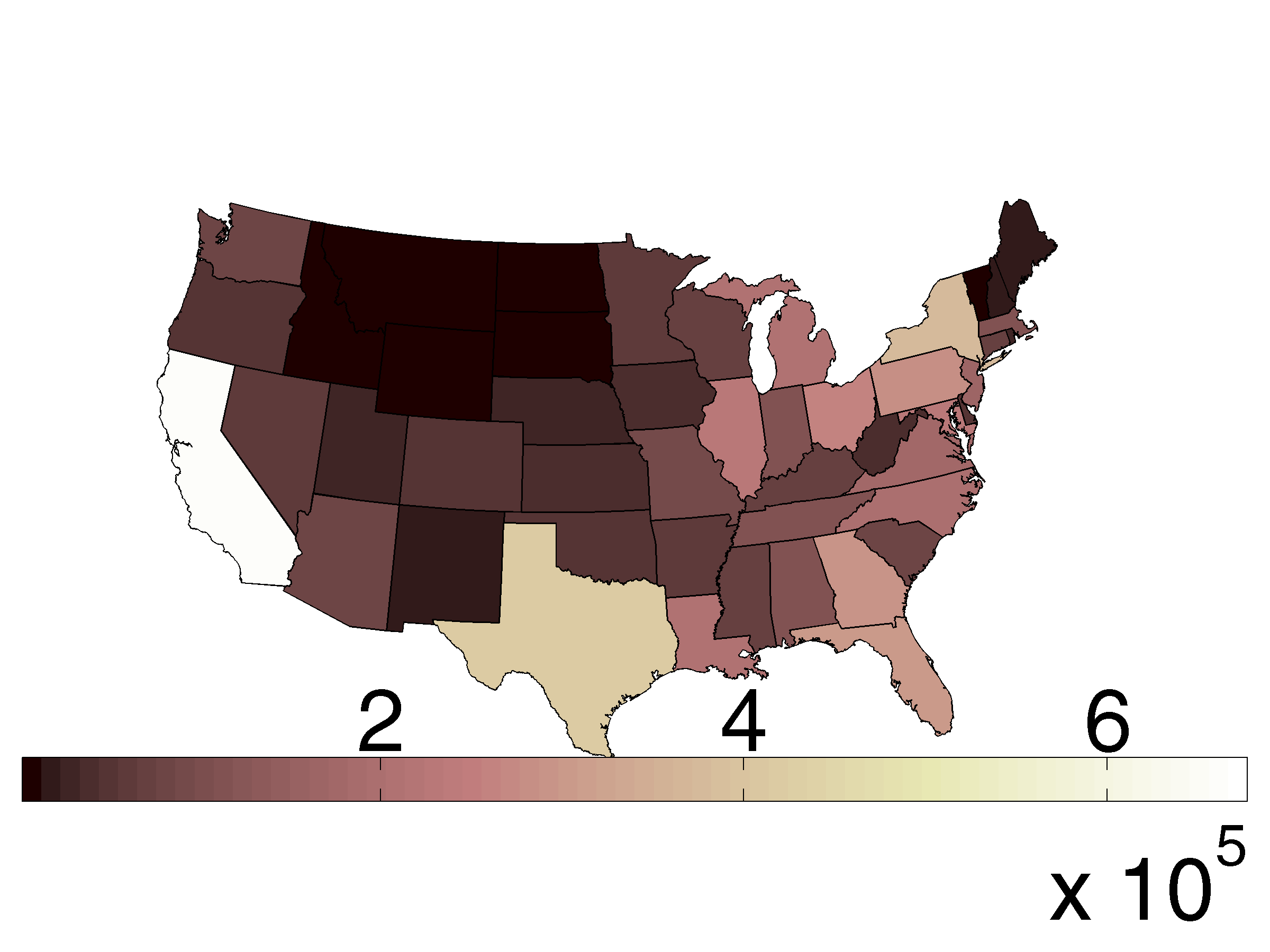} &
\includegraphics[width=.3\columnwidth]{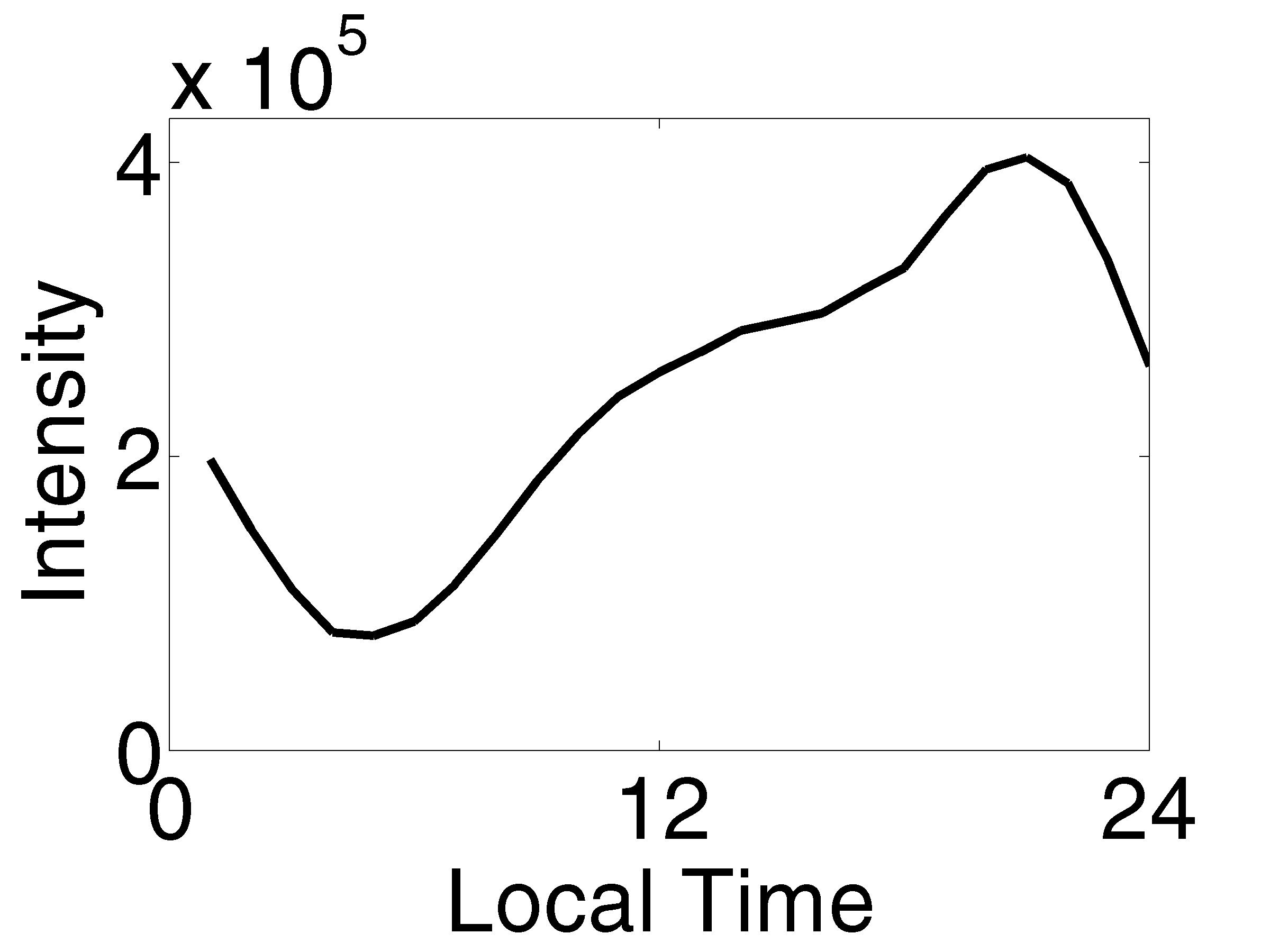} \\
(a) $\widehat{\bg}$ & (b)  spatial & (c) temporal \\
\end{tabular}
\caption{Human population intensity $\widehat{\bg}$.}
\label{fig:g}
\vskip -2ex
\end{figure}

\subsubsection{Overall Counts $\bfz^{(1)}$ and Human Population Intensity $\bg$.}

To obtain the overall counts $\bfz$, we collected tweets through the Twitter stream API using bounding boxes covering continental US. %
The API supplied a subsample of \emph{all} tweets (not just target posts) with geo-tag. 
Therefore, all these tweets include precise latitude and longitude on where they were created.
Through a reverse geocoding database (\url{http://www.datasciencetoolkit.org}), we mapped the coordinates to a US state.
There are a large number of such tweets.
Counting the number of tweets in each state-hour bin gave us $\bfz^{(1)}$, from which $\bg$ is estimated.

Figure~\ref{fig:g} shows the estimated $\widehat{\bg}$.
The x-axis is hour of day and y-axis is the states, ordered by longitude from east (top) to west (bottom).
Although $\widehat\bg$ in this matrix form contains full information, it can be hard to interpret.
Therefore, we visualize aggregated results as well:
First, we aggregate out time in $\widehat\bg$: for each state $s$, we compute $\sum_{t=1}^{24} \widehat{g_{s,t}}$ and show the resulting intensity maps in Figure~\ref{fig:g}(b). 
Second, we aggregate out state in $\widehat\bg$: 
for each hour of day $t$, we compute $\sum_{s=1}^{49} \widehat{g_{s,t}}$ and show the daily curve in Figure~\ref{fig:g}(c).
From these two plots, we clearly see that human population intensity varies greatly both spatially and temporally.

\subsubsection{Identifying Target Posts to Obtain Counts $\bfx$.}

To produce the target counts $\bfx$, we need to first identify target posts describing roadkill events.
Although not part of Socioscope,
we detail this preprocessing step here for reproducibility.

In step 1, we collected tweets using a keyword API.
Each tweet must contain the wildlife name (e.g., ``squirrel(s)'')
\emph{and} the phrase ``ran over''.
We obtained 5857 squirrel tweets, 325 chipmunk tweets, 180 opossum tweets and 159 armadillo tweets during the study period.
However, many such tweets did not actually describe roadkill events.
For example, \emph{``I almost ran over an armadillo on my longboard, luckily my cat-like reflexes saved me.''}
Clearly, the author did not kill the armadillo.

In step 2, we built a binary text classifier to identify target posts among them.
Following \cite{Settles2011Dualist}, the tweets were case-folded without any stemming or stopword removal. 
Any user mentions preceded by a ``@'' were replaced by the anonymized user name ``@USERNAME''.
Any URLs staring with ``http'' were replaced by the token ``HTTPLINK''.
Hashtags (compound words following ``\#'') were not split and were treated as a single token.
Emoticons, such as ``:)'' or ``:D'', were also included as tokens.
Each tweet is then represented by a feature vector consisting of unigram and bigram counts. %
If any unigram or bigram included animal names, we added an additional feature by replacing the animal name with the generic token ``ANIMAL''. For example, we would created an extra feature ``over ANIMAL'' for the bigram ``over raccoon''.
The training data consists of 1,450 manually labeled tweets in August 2011 (i.e., \emph{outside} our study period).  These training tweets contain hundreds of animal species, not just the target species.  The binary label is whether the tweet is a true first-hand roadkill experience.
We trained a linear Support Vector Machine (SVM).  The CV accuracy is nearly 90\%.
We then applied this SVM to classify tweets surviving step 1.
Those tweets receiving a positive label were treated as target posts.

In step 3, we produce $\bfx^{(1)}, \bfx^{(2)}, \bfx^{(3)}$ counts.
Because these target tweets were collected by the keyword API, 
the nature of the Twitter API means that most do not contain precise location information.
As mentioned earlier, only 3\% of them contain coordinates.
We processed this 3\% by the same reverse geocoding database to map them to a US state $s$, and place them in the $x^{(1)}_{s,t}$ detection bins. 
47\% of the target posts do not contain coordinates but can be mapped to a US state from user self-declared profile location.
These are placed in the $x^{(2)}_{s,t}$ detection bins.
The remaining 50\% contained no location meta data, and 
were placed in the $x^{(3)}_{t}$ detection bins.
\footnote{There were actually only a fraction of all tweets without location which came from all over the world.
We estimated this US/World fraction using $\bfz$.}

\subsubsection{Constructing the Transition Matrix $\bP$.}
In this study, $\bP$ characterizes the fraction of tweets which were actually generated in source bin $(s,t)$ end up in the three detector bins: precise location ${st}^{(1)}$, potentially noisy location ${st}^{(2)}$, and missing location ${t}^{(3)}$.
We define $\bP$ as follows:

$\bP_{{(s,t)}^{(1)}, (s,t)}=0.03$, and $\bP_{{(r,t)}^{(1)}, (s,t)}=0$ for $\forall r \neq s$ to reflect the fact that we know precisely 3\% of the target posts' location.

$\bP_{{(r,t)}^{(2)}, (s,t)}=0.47 M_{r,s}$ for all $r,s$. $M$ is a $49 \times 49$ ``mis-self-declare'' matrix.
$M_{r,s}$ is the probability
that a user self-declares in her profile that she is in state $r$, but her post is in fact generated in state $s$.
We estimated $M$ from a separate large set of tweets with both coordinates and self-declared profile locations.
The $M$ matrix is asymmetric and interesting in its own right:
many posts self-declared in California or New York were actually produced all over the country; 
many self-declared in Washington DC were actually produced in Maryland or Virgina;
more posts self-declare Wisconsin but were actually in Illinois than the other way around.

$\bP_{t^{(3)}, (s,t)}=0.50$. This aggregates tweets with missing information into the third kind of detector bins.

\subsubsection{Specifying the Graph Regularizer.}
Our graph has two kinds of edges.
Temporal edges connect source bins with the same state and adjacent hours by weight $w_t$.
Spatial edges connect source bins with the same hour and adjacent states by weight $w_s$.
The regularization weight $\lambda$ was absorbed into $w_t$ and $w_s$.
We tuned the weights $w_t$ and $w_s$ with CV on the 2D grid $\{10^{-3}, 10^{-2.5}, \ldots, 10^3\}^2$.

\subsection{Results}

\begin{figure}[t!]
\centering
\begin{tabular}{ccc}
\includegraphics[width=.3\columnwidth]{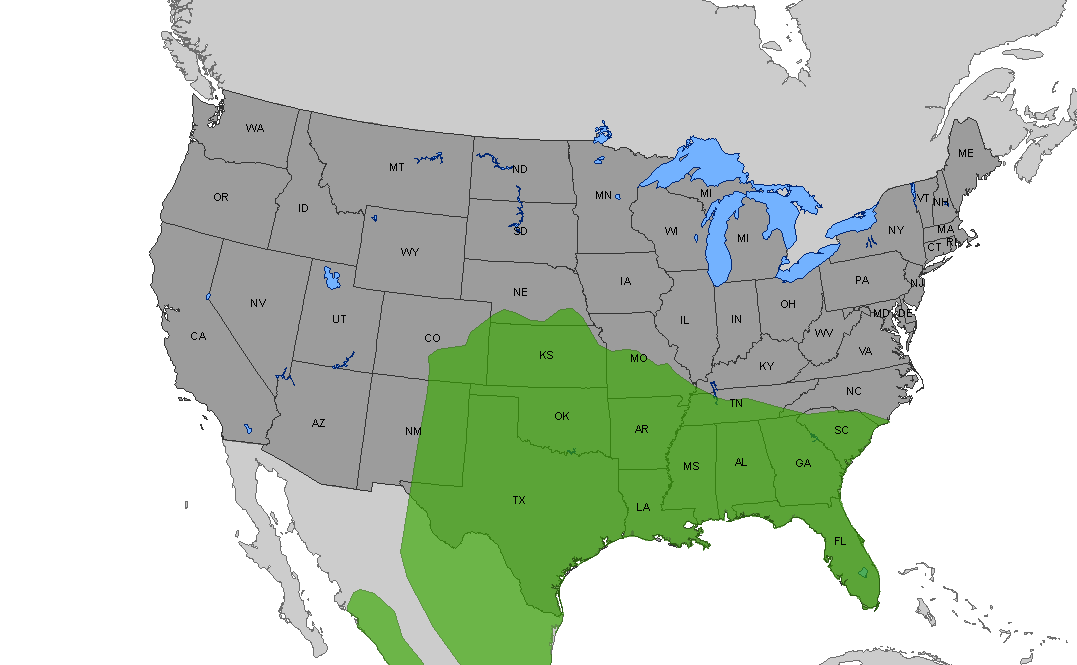} &
\includegraphics[width=.3\columnwidth]{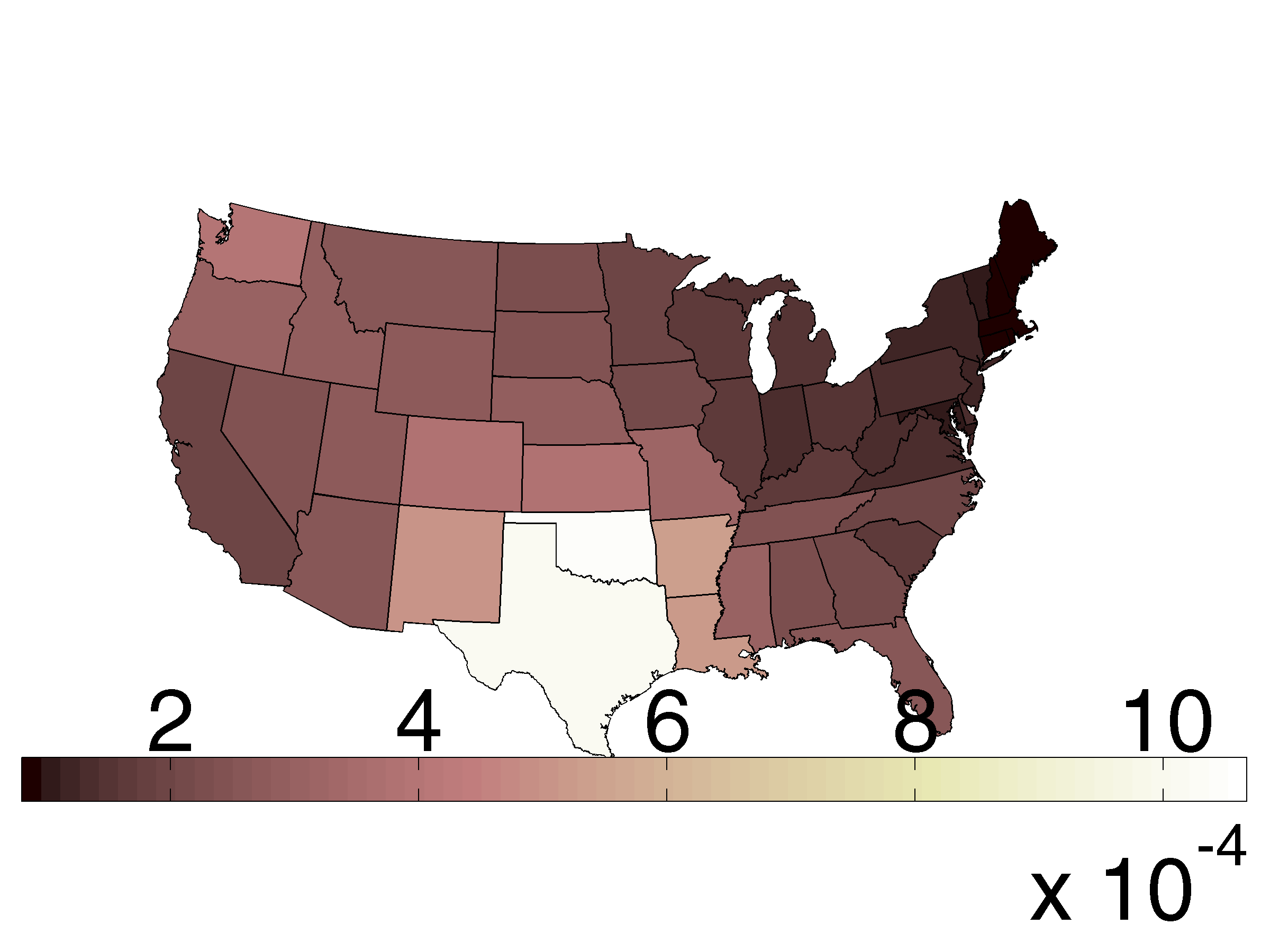} &
\includegraphics[width=.3\columnwidth]{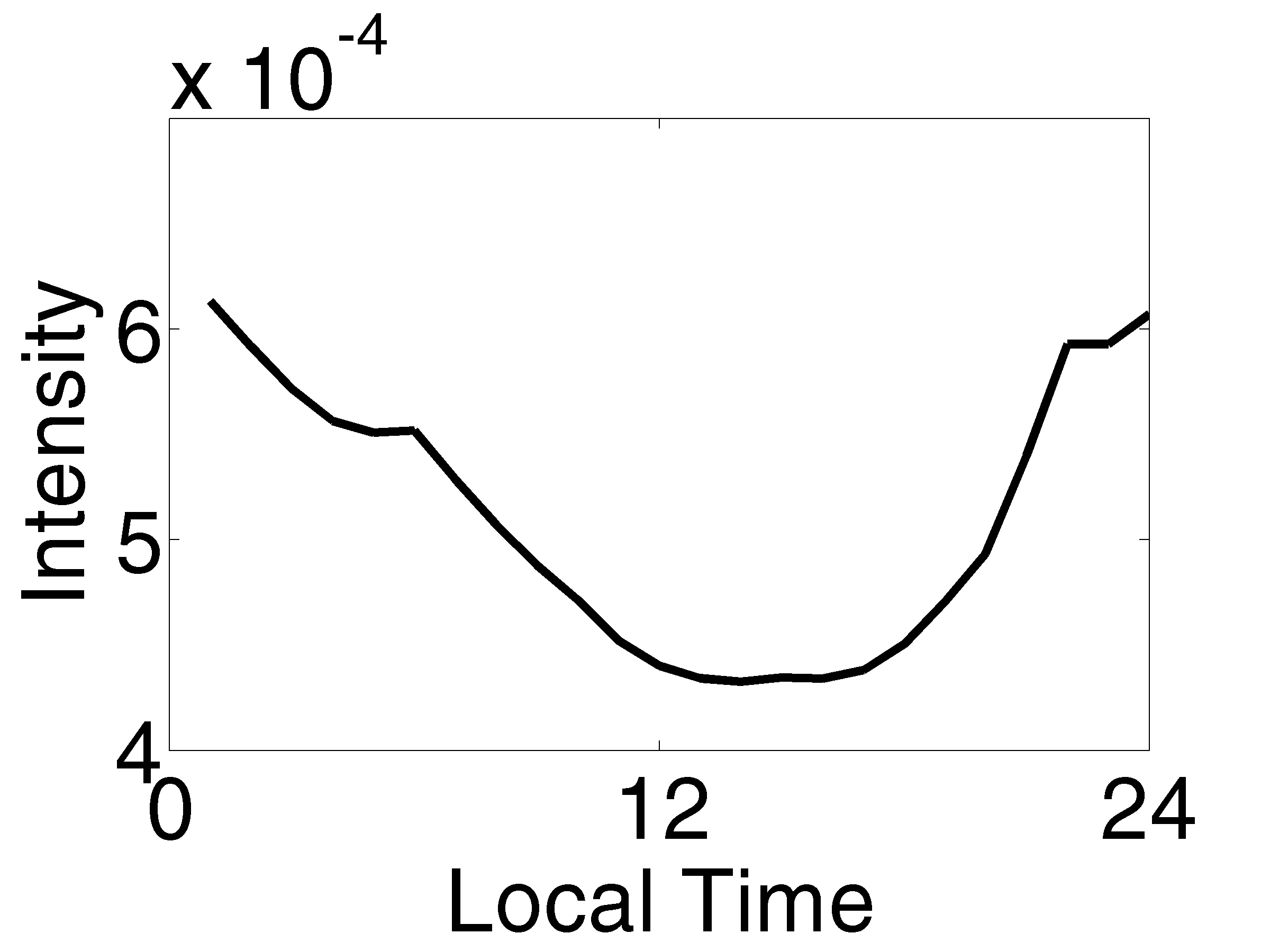} \\
\multicolumn{3}{c}{(a) armadillo (Dasypus novemcinctus)} \\

\includegraphics[width=.3\columnwidth]{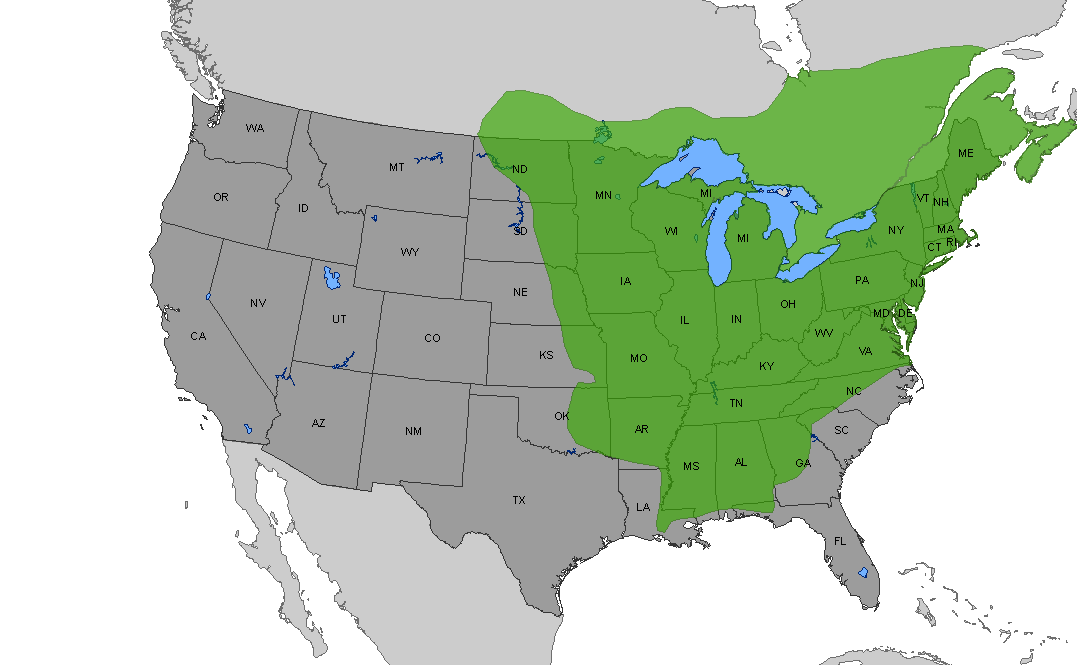} &
\includegraphics[width=.3\columnwidth]{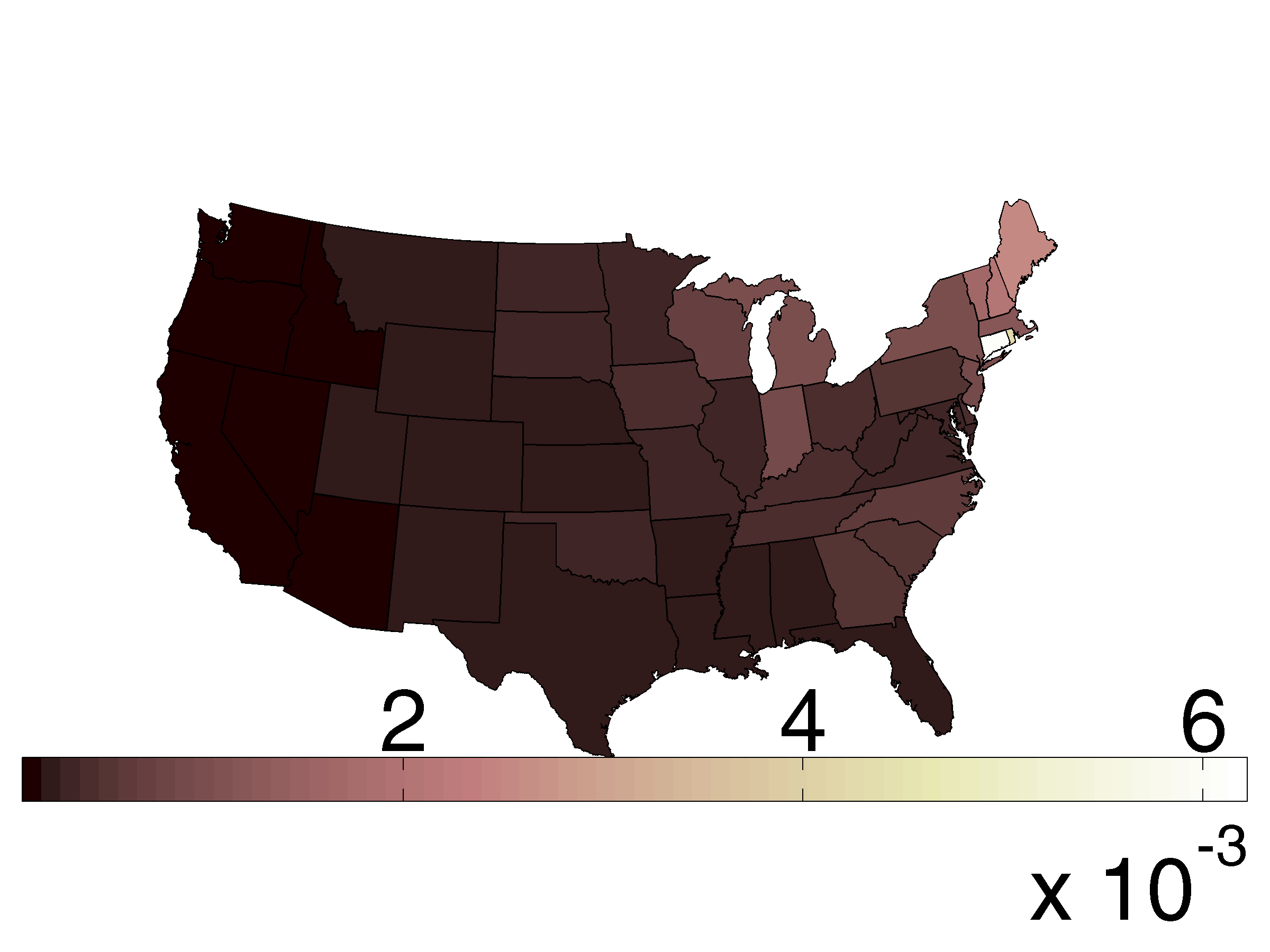} &
\includegraphics[width=.3\columnwidth]{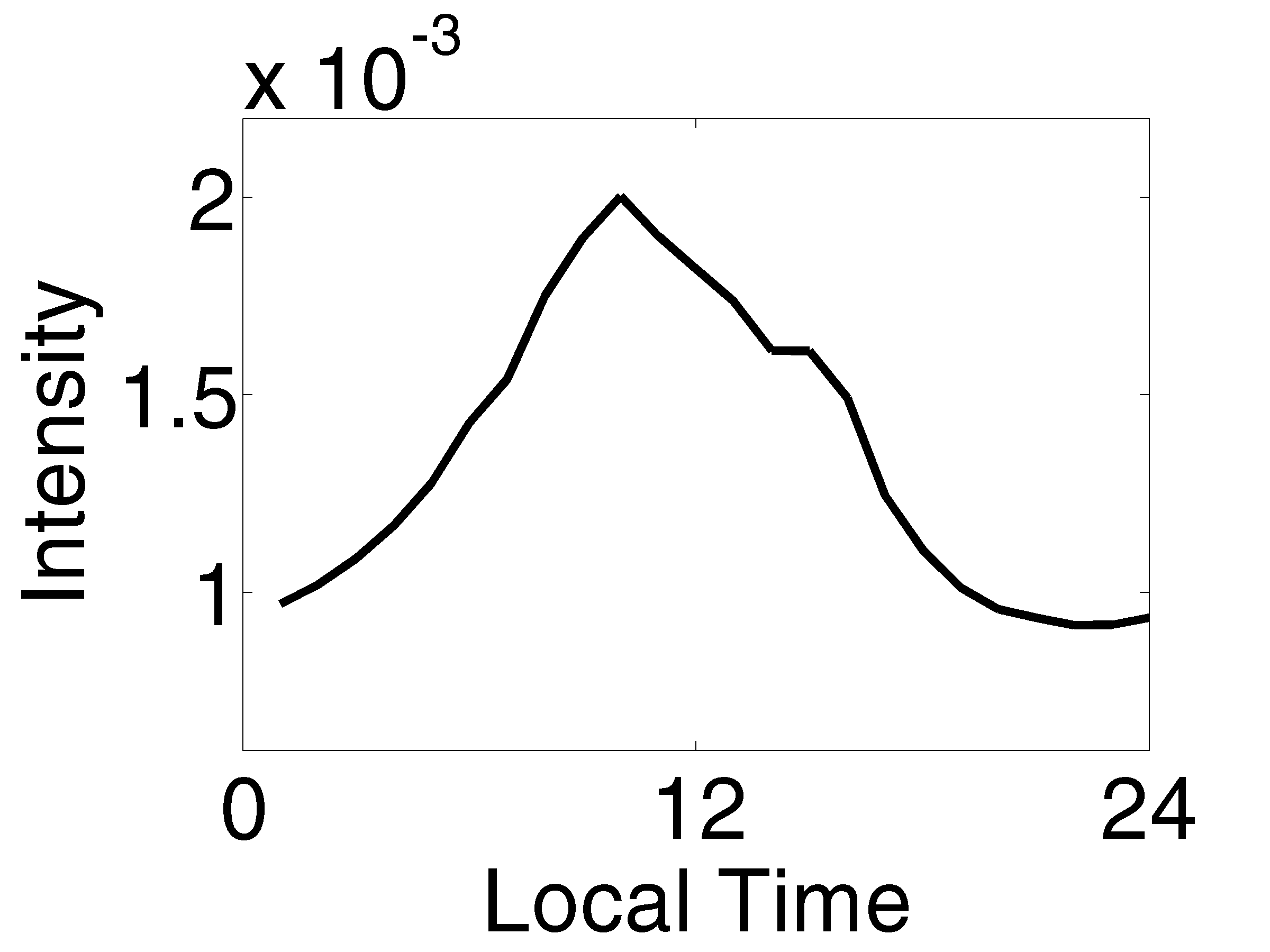} \\
\multicolumn{3}{c}{(b) chipmunk (Tamias striatus)} \\

\includegraphics[width=.3\columnwidth]{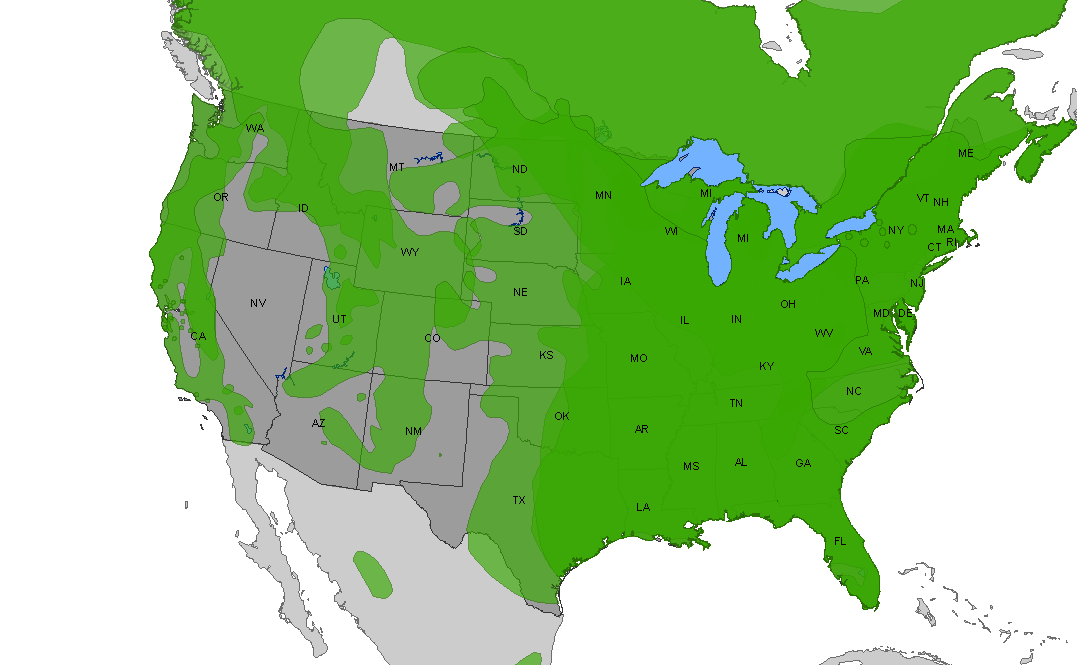} &
\includegraphics[width=.3\columnwidth]{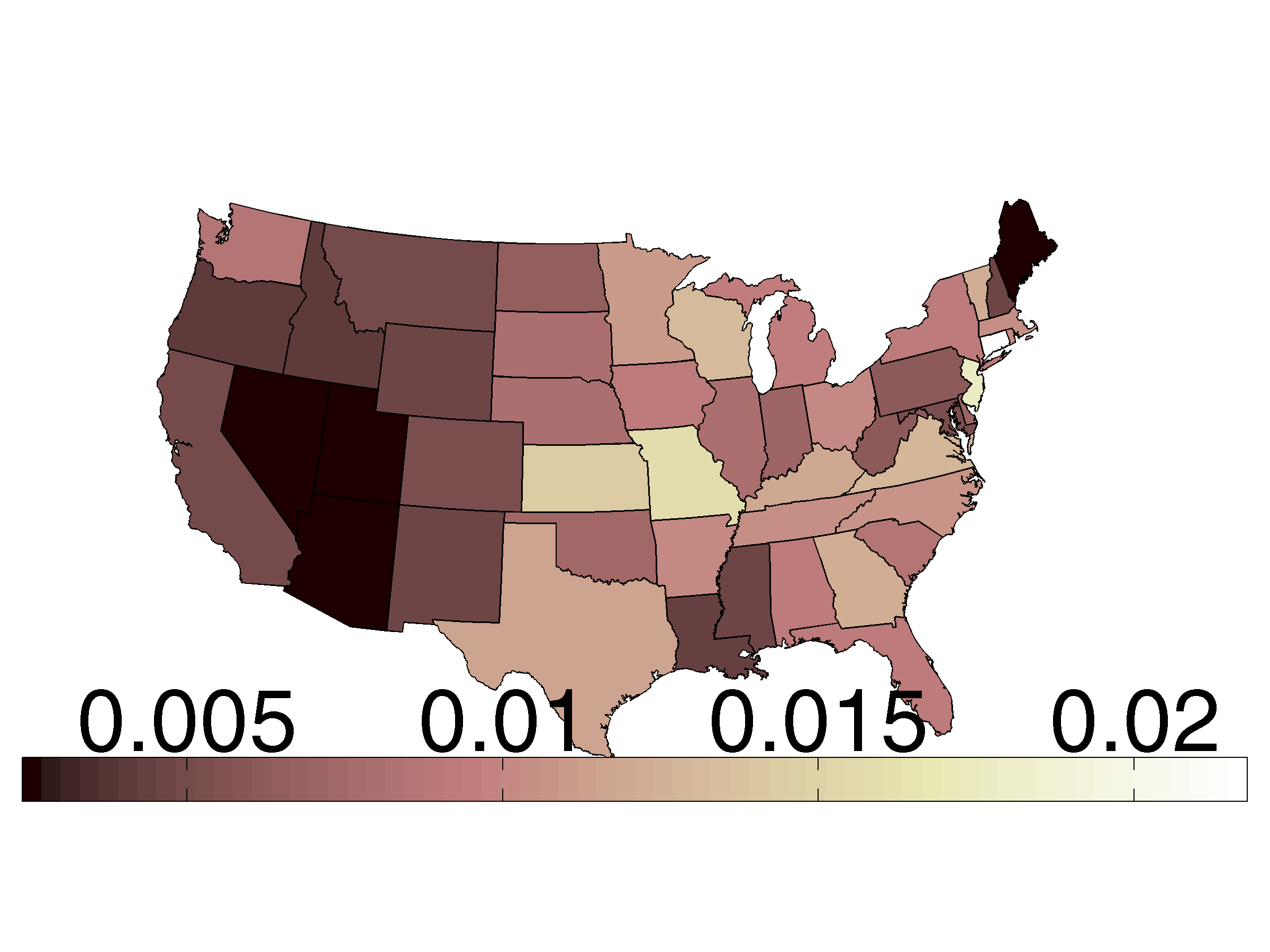} &
\includegraphics[width=.3\columnwidth]{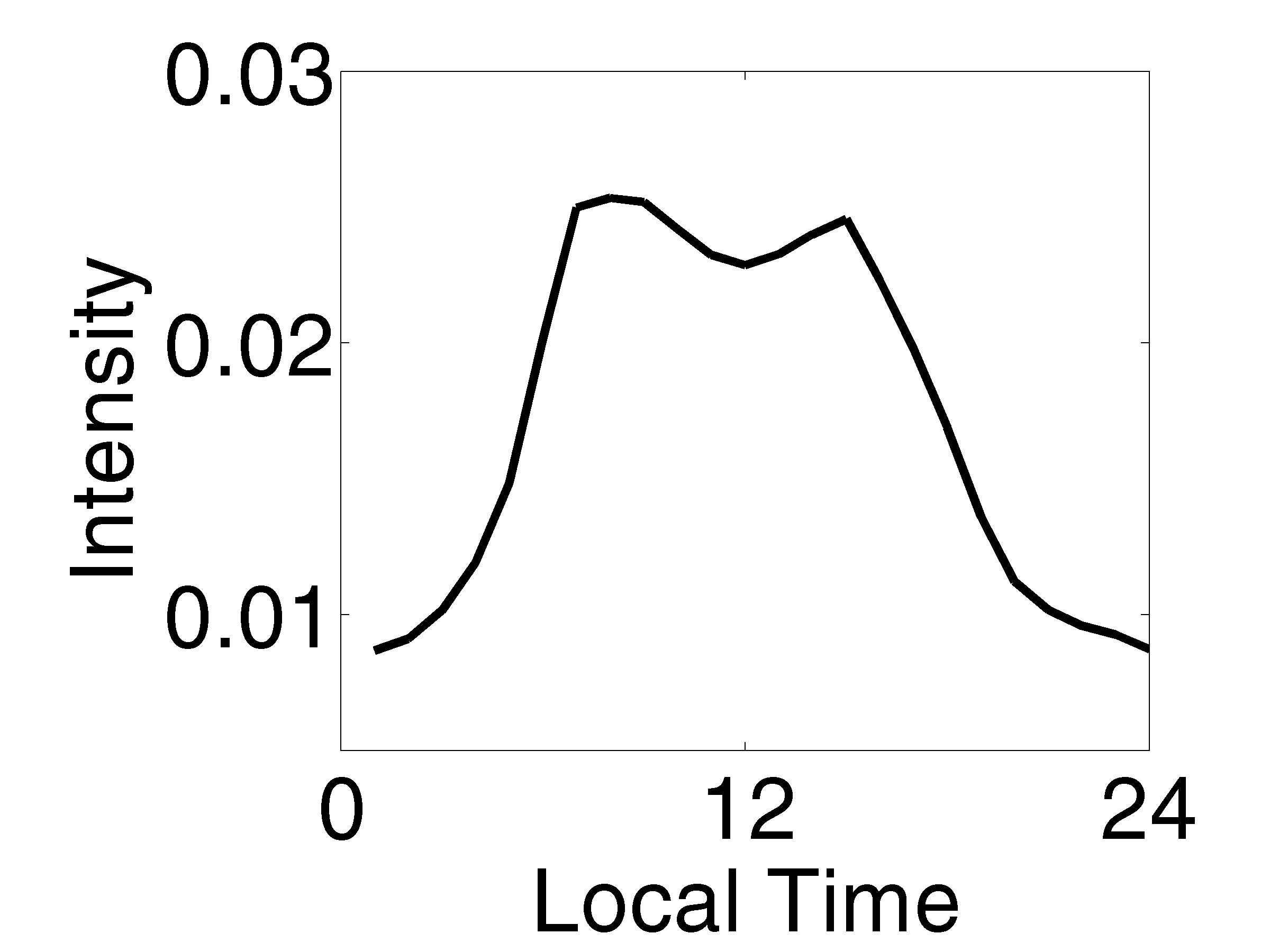} \\
\multicolumn{3}{c}{(c) squirrel (Sciurus carolinensis and several others)}\\

\includegraphics[width=.3\columnwidth]{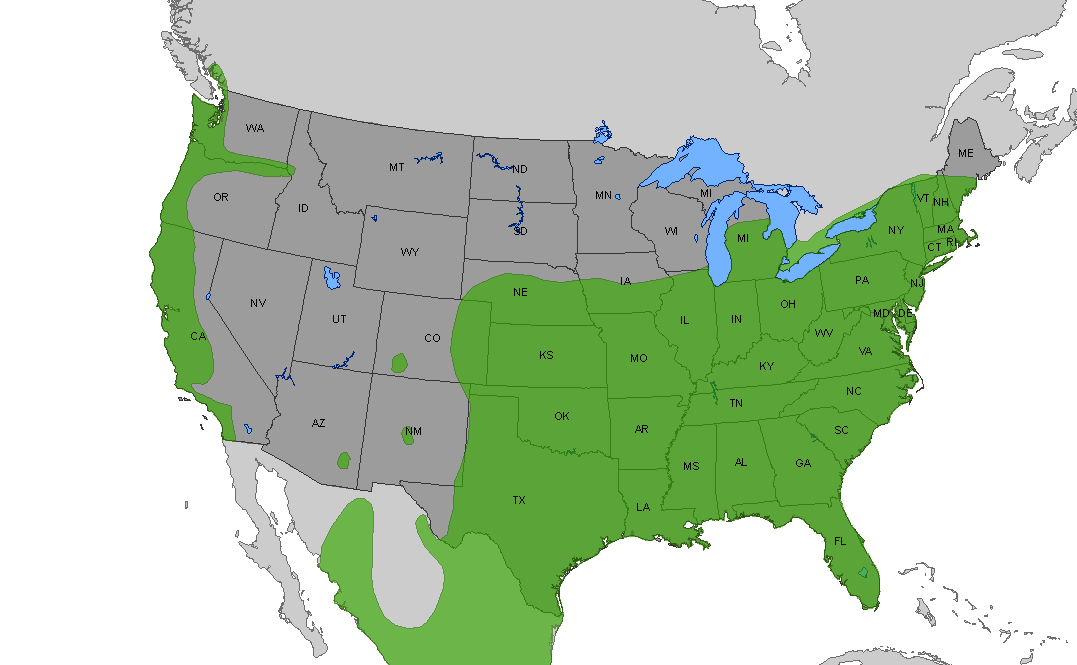} &
\includegraphics[width=.3\columnwidth]{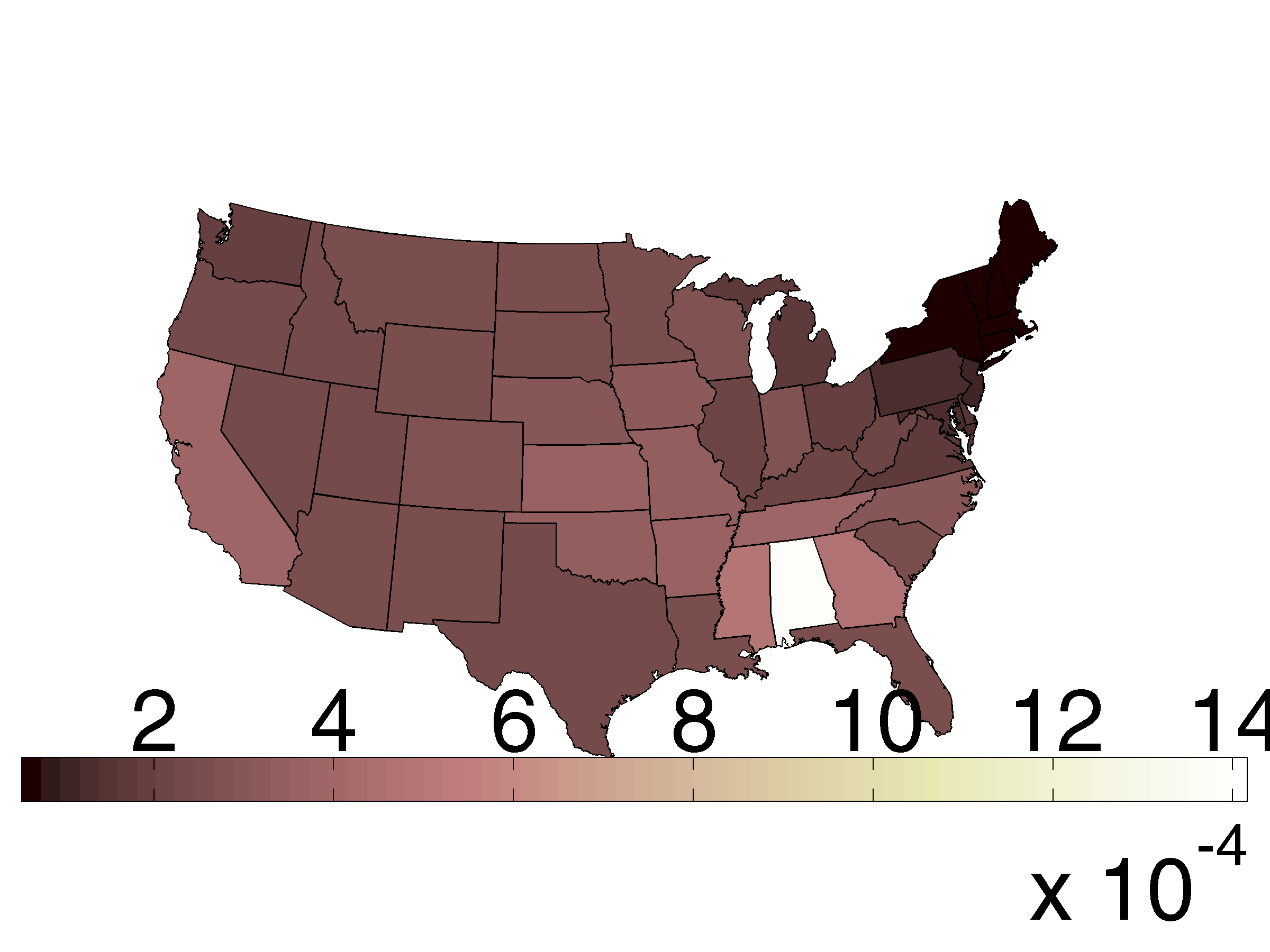} &
\includegraphics[width=.3\columnwidth]{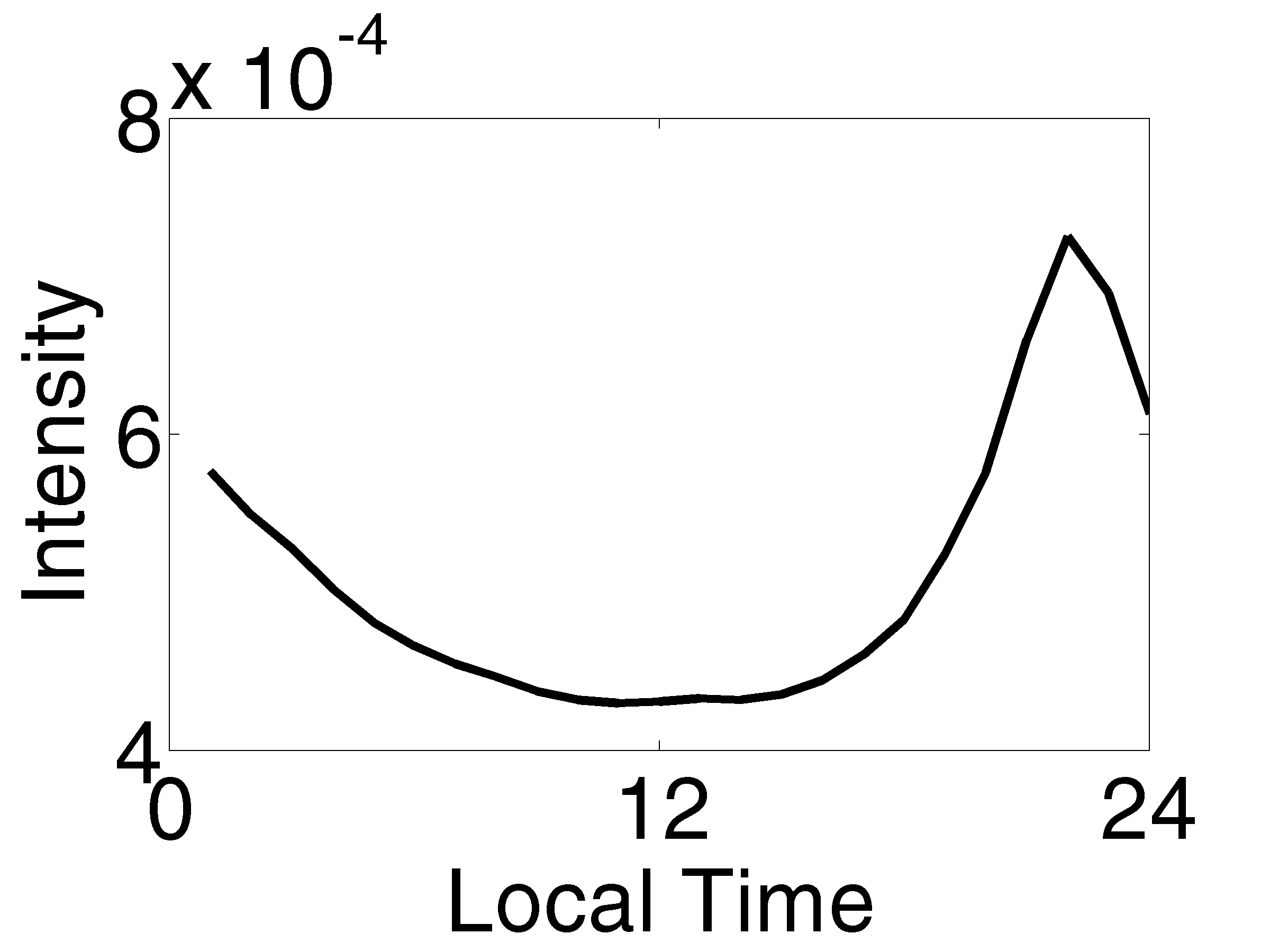} \\
\multicolumn{3}{c}{(d) opossum (Didelphis virginiana)} \\
\end{tabular}
\caption{Socioscope estimates match animal habits well. (Left) range map from NatureServe, 
(Middle) Socioscope $\widehat\bff$ aggregated spatially,
(Right) $\widehat\bff$ aggregated temporally.
} 
\label{fig:four}
\vskip -2ex
\end{figure}

We present results on four animals: armadillos, chipmunks, squirrels, opossums.
Perhaps surprisingly, precise roadkill intensities for these animals are apparently unknown to science
(This serves as a good example of the value Socioscope may provide to wildlife scientists).
Instead, domain experts were only able to provide a range map of each animal, see the left column in Figure~\ref{fig:four}.
These maps indicate presence/absence only, and were extracted from NatureServe~\cite{natureServe2007}.
In addition, the experts defined 
armadillo and opossum as nocturnal, 
chipmunk as diurnal, and squirrels as both crepuscular (active primarily during twilight) and diurnal.
Due to the lack of quantitative ground-truth, our comparison will necessarily be qualitative in nature.

\begin{figure}[t!]
\centering
\begin{tabular}{cc}
\includegraphics[width=.3\columnwidth]{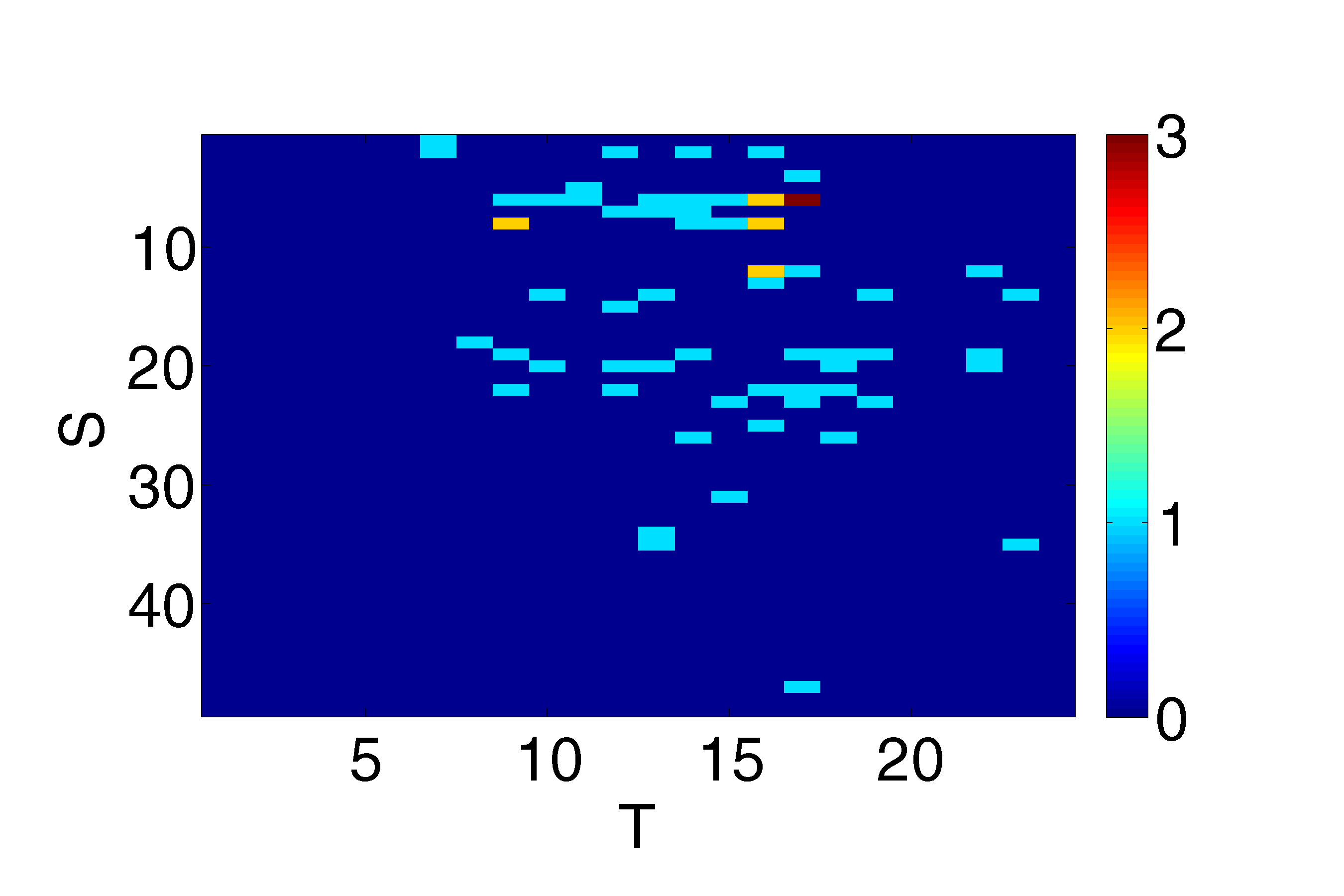} &
\includegraphics[width=.3\columnwidth]{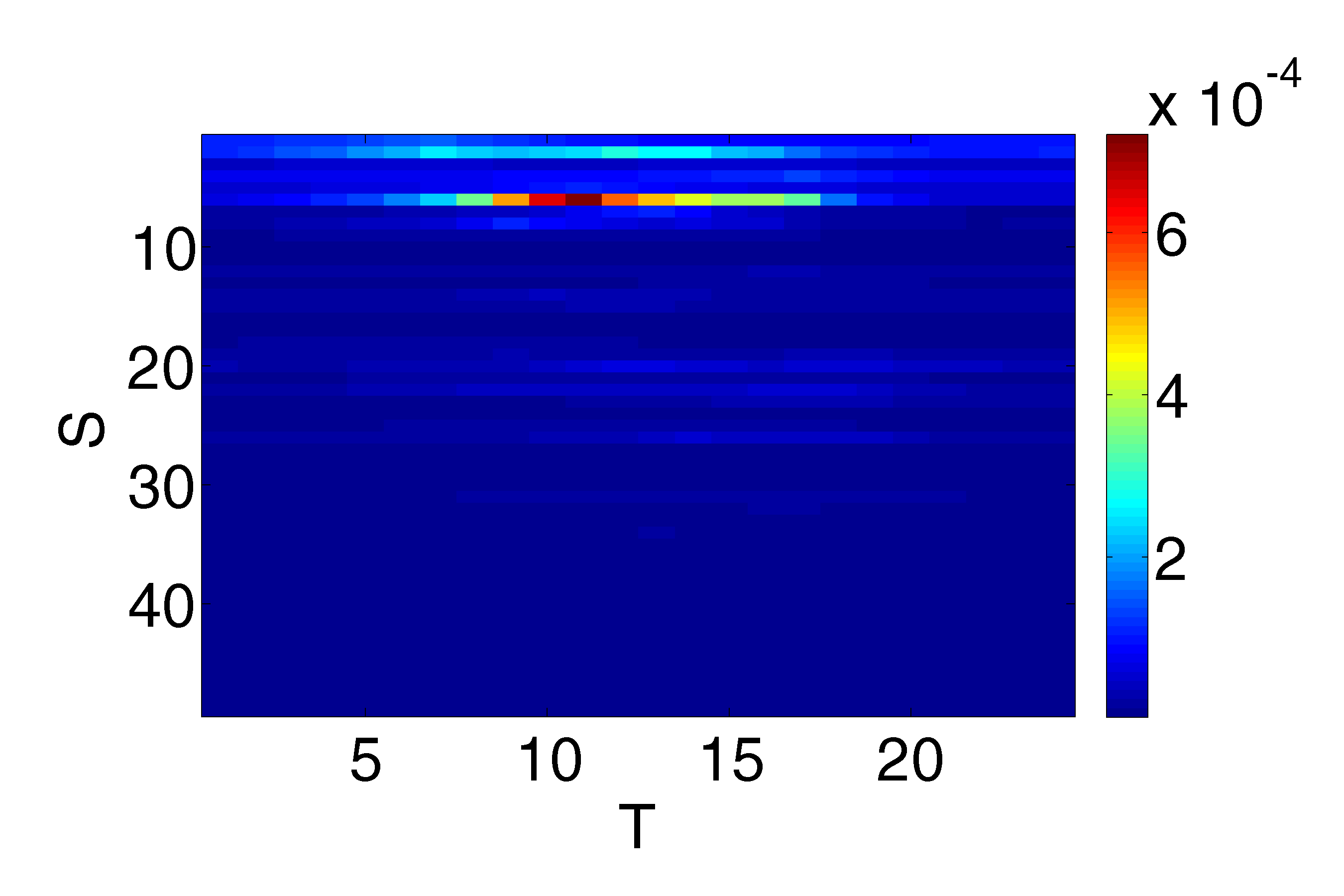}  \\
(a) $\bfx^{(1)} + \bfx^{(2)}$ & (b)  Socioscope $\widehat\bff$ \\
\end{tabular}
\caption{Raw counts and Socioscope $\widehat\bff$ for chipmunks}
\label{fig:matrix}
\vskip -2ex
\end{figure}

Socioscope provides sensible estimates on these animals.
For example, Figure~\ref{fig:matrix}(a) shows counts $\bfx^{(1)}+\bfx^{(2)}$ for chipmunks which is very sparse (the largest count in any bin is 3), and Figure~\ref{fig:matrix}(b) the Socioscope estimate $\widehat{\bff}$.
The axes are the same as in Figure~\ref{fig:g}(a).
In addition, we present the state-by-state intensity maps in the middle column of Figure~\ref{fig:four} by aggregating $\widehat\bff$ spatially.
The Socioscope results match the range maps well for all animals.
The right column in Figure~\ref{fig:four} shows the daily animal activities by aggregating $\widehat\bff$ temporally.
These curves match the animals' diurnal patterns well, too.

\begin{figure}[t!]
\centering
\begin{tabular}{ccc}
\includegraphics[width=.3\columnwidth]{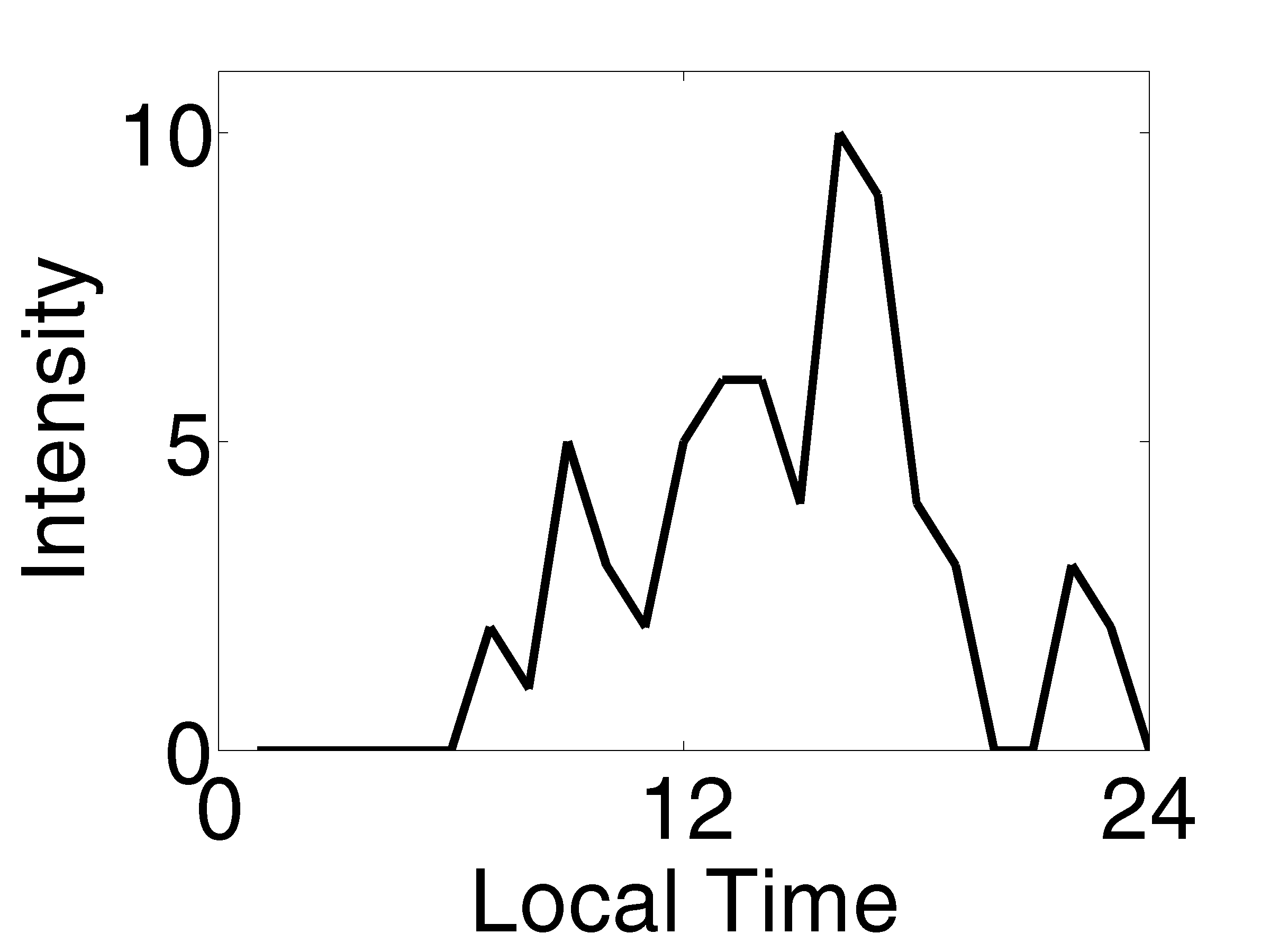} &
\includegraphics[width=.3\columnwidth]{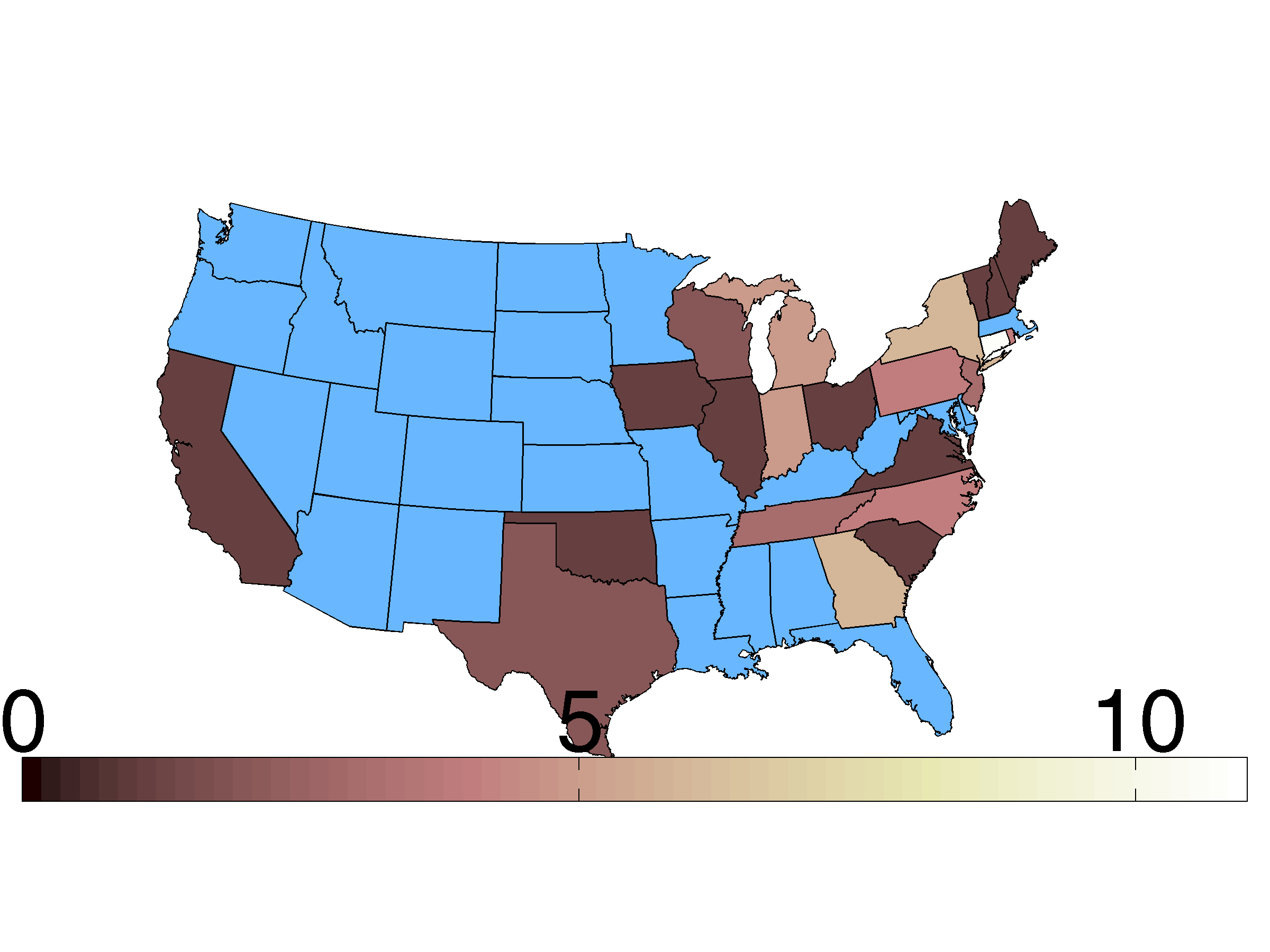} &
\includegraphics[width=.3\columnwidth]{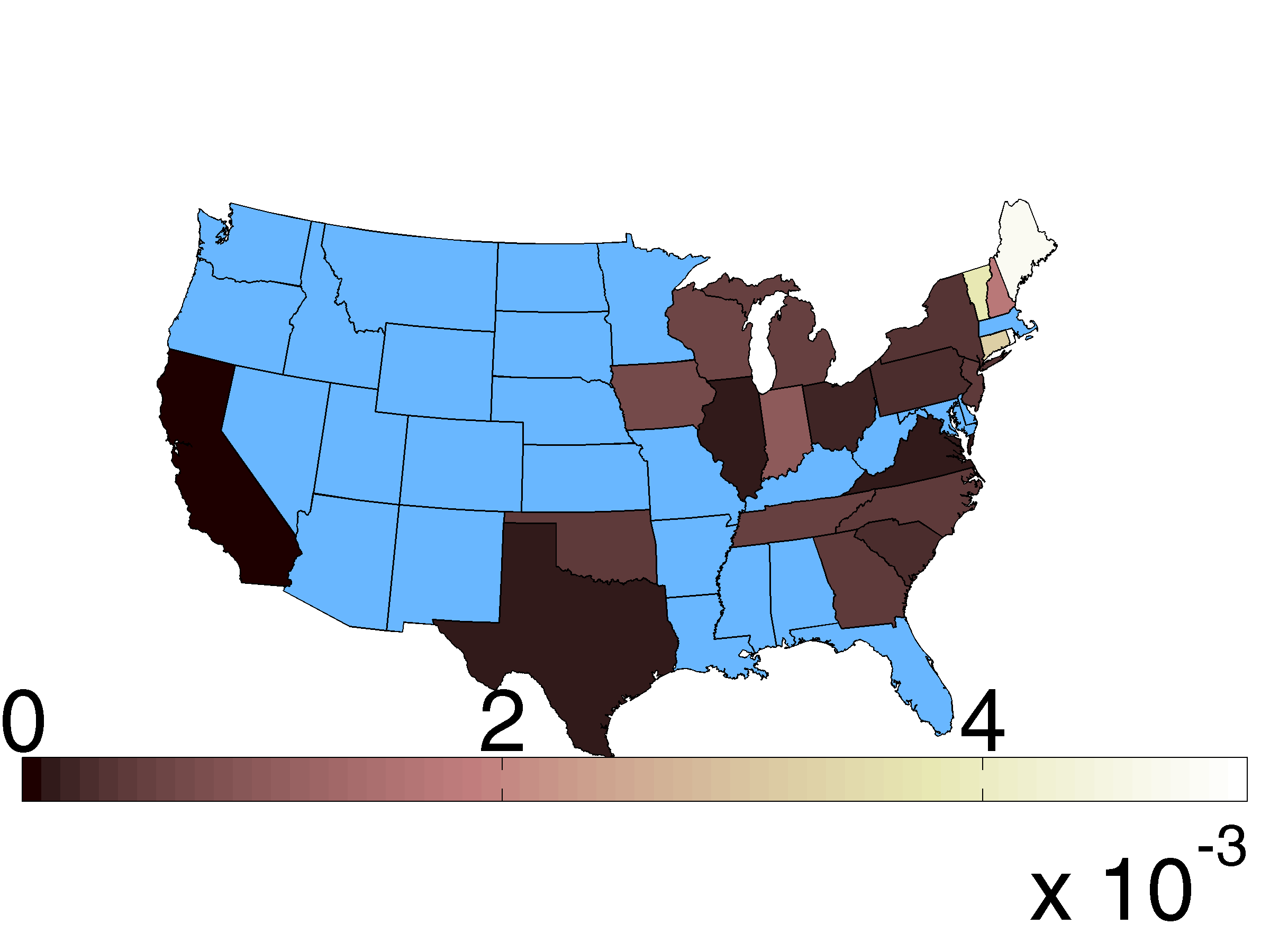} \\
(a) $\bfx^{(1)} + \bfx^{(2)}$  & (b) $\bfx^{(1)} + \bfx^{(2)}$  & (c) $(\bfx^{(1)} + \bfx^{(2)})/ \bfz^{(1)}$ \\
\end{tabular}
\caption{Examples of inferior baseline estimators. In all plots, states with zero counts are colored in blue.}
\label{fig:simple}
\vskip -2ex
\end{figure}

The Socioscope estimates are superior to the baseline methods in Table~\ref{tab:mse}.
Due to space limit we only present two examples on chipmunks, but note that similar observations exist for all animals.
The baseline estimator of simply scaling $\bfx^{(1)}+\bfx^{(2)}$ produced the temporal and spatial aggregates in Figure~\ref{fig:simple}(a,b).
Compared to Figure~\ref{fig:four}(b, right), the temporal curve has a spurious peak around 4-5pm.
The spatial map contains spurious intensity in California and Texas, states outside the chipmunk range as shown in Figure~\ref{fig:four}(b, left).
Both are produced by population bias when and where there were strong background social media activities (see Figure~\ref{fig:g}(b,c)).
In addition, the spatial map contains 27 ``holes'' (states with zero intensity, colored in blue)  due to data scarcity.
In contrast, Socioscope's estimates in Figure~\ref{fig:four} avoid this problem by regularization.
Another baseline estimator $(\bfx^{(1)}+\bfx^{(2)})/\bfz^{(1)}$ is shown in Figure~\ref{fig:simple}(c). 
Although corrected for population bias, this estimator lacks the transition model and regularization. 
It does not address data scarcity either.

%% file: conclusion.tex
\section{Future Work}

Using social media as a data source for spatio-temporal signal recovery is an emerging area.
Socioscope represents a first step toward this goal.
There are many open questions: 

1. We treated target posts as certain.
In reality, a natural language processing system can often supply a confidence.
For example, a tweet might be deemed to be a target post only with probability 0.8.
It will be interesting to study ways to incorporate such confidence into our framework.

2. The temporal delay and spatial displacement between the target event and the generation of a post is commonplace, as discussed in footnote~\ref{fn:psf}.
Estimating an appropriate transition matrix $P$ from social media data so that Socioscope can handle such ``point spread functions'' remains future work.

3. It might be necessary to include psychology factors to better model the human ``sensors.''
For instance, a person may not bother to tweet about a chipmunk roadkill, but may be eager to do so upon seeing a moose roadkill.

4. Instead of discretizing space and time into bins, one may adopt a spatial point process model to learn a continuous intensity function instead~\cite{moller2004statistical}.

Addressing these considerations will further improve Socioscope.